\documentclass[a4paper]{svproc}

\usepackage{url}

\usepackage[noend]{algorithmic}
\usepackage{comment}
\usepackage{float}
\usepackage{graphicx}
\usepackage[section]{placeins}
\usepackage{tikz}
\usetikzlibrary{math,positioning,calc,fit,arrows,matrix}
\usepackage[ruled,linesnumbered,vlined]{algorithm2e}
\usepackage{subcaption}
\usepackage{multirow}
\usepackage{array}

\begin{document}
\mainmatter
\title{ Incorporating Stochastic Models of Controller Behavior into Kinodynamic Efficiently Adaptive State Lattices for Mobile Robot Motion Planning in Off-Road Environments}
\titlerunning{Incorporating Stochastic Models of Controller Response into KEASL}  
%
\author{Eric R. Damm\inst{1} \and Eli S. Lancaster\inst{2}
\and Felix A. Sanchez\inst{3}\and\\ Kiana Bronder\inst{3}
\and Jason M. Gregory\inst{2} \and Thomas M. Howard\inst{1,2}}
\authorrunning{Eric R. Damm et al.} 
%
\tocauthor{Eric R. Damm, Eli S. Lancaster, Felix A. Sanchez, Kiana Bronder, Jason M. Gregory, Thomas M. Howard}
\institute{University of Rochester, Rochester NY, USA,\\
\email{edamm@ur.rochester.edu},
\and
DEVCOM Army Research Laboratory, Adelphi MD, USA, \\
\and
Parsons, Chantilly VA, USA
}

\maketitle              
 
\begin{abstract}
	Mobile robot motion planners rely on theoretical models to predict how the robot will move through the world.
	However, when deployed on a physical robot, these models are subject to errors due to real-world physics and uncertainty in how the lower-level controller follows the planned trajectory.
	In this work, we address this problem by presenting three methods of incorporating stochastic controller behavior into the recombinant search space of the Kinodynamic Efficiently Adaptive State Lattice (KEASL) planner.
	To demonstrate this work, we analyze the results of experiments performed on a Clearpath Robotics Warthog Unmanned Ground Vehicle (UGV) in an off-road, unstructured environment using two different perception algorithms, and performed an ablation study using a full spectrum of simulated environment map complexities.
	Analysis of the data found that incorporating stochastic controller sampling into KEASL leads to more conservative trajectories that decrease predicted collision likelihood when compared to KEASL without sampling.
	When compared to baseline planning with expanded obstacle footprints, the predicted likelihood of collisions becomes more comparable, but reduces the planning success rate for baseline search.
	\keywords{kinodynamic motion planning, off-road robotics, unstructured environments, stochastic uncertainty, robot navigation, UGV}
\end{abstract}

\section{Introduction}\label{sec:introduction}
\vspace{-.3cm}
Robot navigation software architectures vary, but a common approach involves a hierarchical structure: a high-level motion planner generates trajectories, which a lower-level path-following controller executes by sending velocity commands to the robot's actuators.
The motion planner relies on theoretical models to predict how the robot will move through the world.
In practice, real-world physics and uncertainty about controller behavior lead to errors in the model assumptions.
Typically, these sources of error are decoupled, and the planner accounts for inaccuracies in its motion model, while the controller manages deviations from the planned path.
This separation abstracts the controller's realized motion from the planner, potentially leading to incorrect assumptions about what can be executed, resulting in unintended behavior.

In this work, we address the problem of incorporating stochastic controller behavior into the recombinant search space of a high-level planner.
The approaches described herein build upon the Kinodynamic Efficiently Adaptive State Lattice (KEASL) \cite{damm2023terrain}.
KEASL uses speed, elevation, and obstacle maps to compute minimum-cost trajectories via estimated traversal time (also referred to as ``path duration'').
In this work we augment the KEASL search space with stochastic predictions of controller behavior, prompting more conservative plans through unstructured environments.
We explore three methods of doing so.
The first method uses stochastic controller sampling for every node expansion, and marks the node invalid if any control motions are deemed in collision with a lethal obstacle.
The second approach incorporates a node-revision step after any node expands to the goal, and only performs controller sampling from the start state to the candidate final node (instead of every expansion).
The final approach uses the same node-revision step, and also revises the graph based on the node corresponding to a rollout collision earlier in the planned trajectory.

We examine the practicality of each method for on-robot use through data collected from experiments on a Clearpath Robotics Warthog Unmanned Ground Vehicle (UGV) in an off-road, unstructured environment.
The field experiments were performed with two different perception systems: The first (herein referred to as ``Mapping Algorithm 1'') was a neural network based approach from \cite{meng2023terrainnetvisualmodelingcomplex}, and the second (herein referred to as ``Mapping Algorithm 2'') was a geometric approach where occupied voxels detected at a certain height threshold are marked as obstacles.
To more fully analyze the approach, we also include an ablation study using a wide variety of simulated maps generated with Perlin noise.
We analyze the resulting plans relative to those generated by a controller-agnostic version of the same search process.
\vspace{-.3cm}
\section{Related Work}\label{sec:relatedWork}
\vspace{-.3cm}
There are many examples of modeling uncertainty and controller behavior at different levels of a mobile robot's navigation architecture.
Linear Quadratic Gaussian (LQG) control \cite{athans1971role} is a closed-loop control method for predicting a system's performance under physical uncertainty.
This type of control was used to estimate the collision probabilities (CP) of optimal trajectories generated with Monte Carlo Motion Planning (MCMP) \cite{janson2017monte}.
After the CP was calculated, obstacle expansions in C-Space were adjusted to bring the CP to within a desired threshold for the next iteration of optimal trajectory generation.
Belief space planning \cite{bry2011rapidly} is another method that plans motions based on a robot's ability to localize at each node, but assumes accurate motion execution by the lower level controller.
Similarly, \cite{gonzalez2014state} incorporate an uncertainty prediction into the motion primitives of a state lattice, and update the beliefs using an Extended Kalman Filter (EKF) based on sensor uncertainty and proximity to landmarks.
Earlier work using particle-based uncertainty sampling was incorporated into Rapidly-exploring Random Trees (RRT) \cite{melchior2007particle}.
Planning in the belief space prioritizes motions that provide high state certainty rather than what traditional search would consider to be the most efficient \cite{prentice2009belief}.
Other works attempt to solve this problem by assuming motion deviation responsibility to the path following controller by incorporating slip-aware dynamics models into model predictions \cite{williams2016aggressive,rajagopalan2016slip}.
\vspace{-.3cm}
\section{Technical Approach}\label{sec:technicalApproach}
\begin{figure}[htb]
	\begin{center}
	\setlength{\fboxsep}{0pt} 
  	\setlength{\fboxrule}{1pt}
	\begin{subfigure}{0.24\textwidth}
		\fbox{\includegraphics[height=\textwidth,angle=-90]{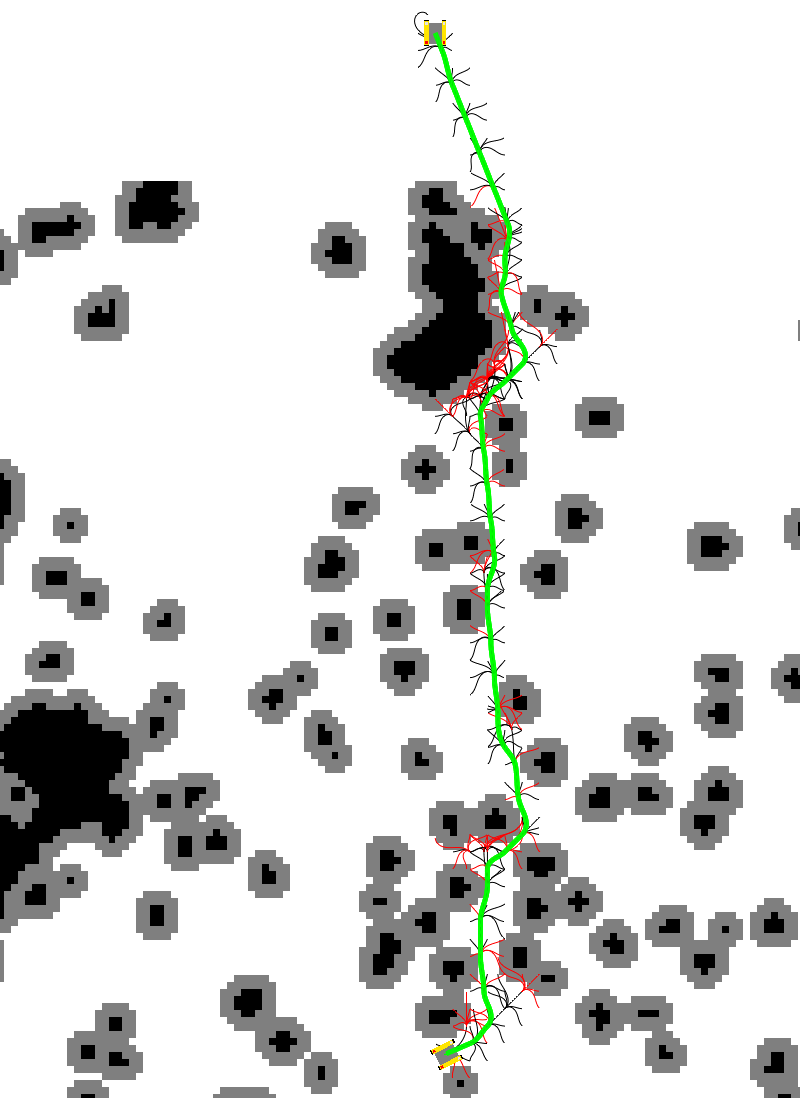}}
		\caption{}\label{fig:easlOne}
	\end{subfigure}
	\begin{subfigure}{0.24\textwidth}
		\fbox{\includegraphics[height=\textwidth,angle=-90]{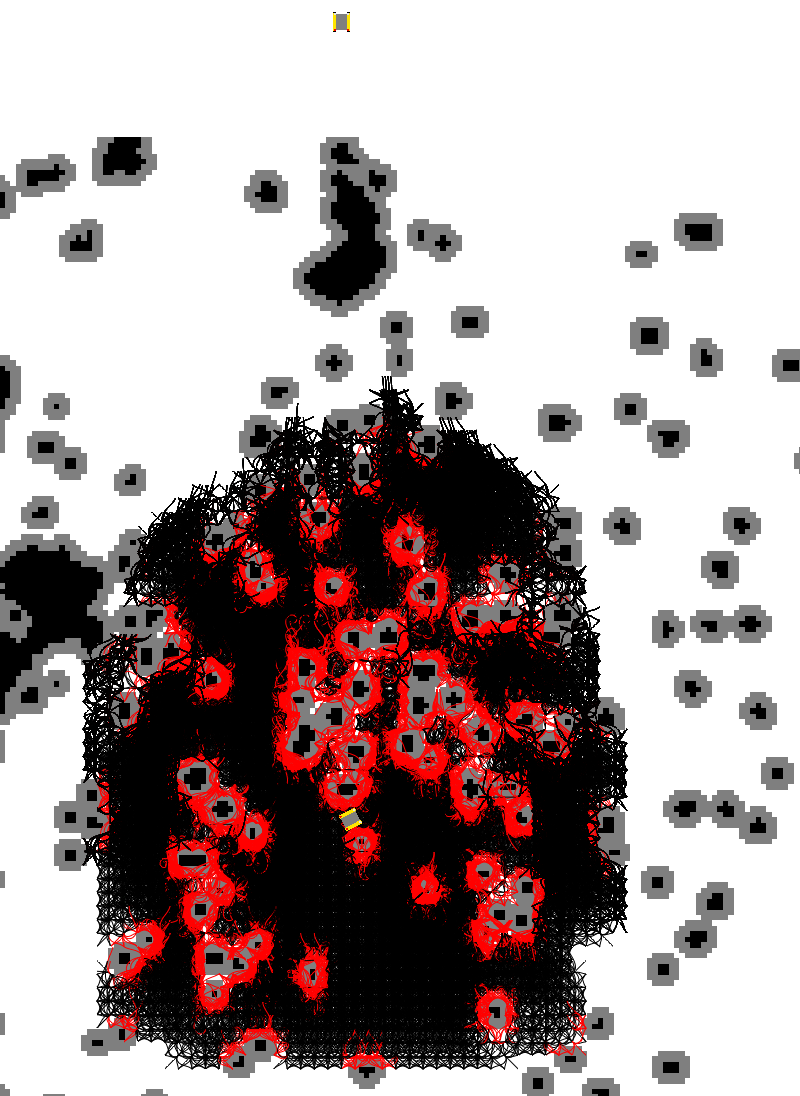}}
		\caption{}\label{fig:methodOne}
	\end{subfigure}
	\begin{subfigure}{0.24\textwidth}
		\fbox{\includegraphics[height=\textwidth,angle=-90]{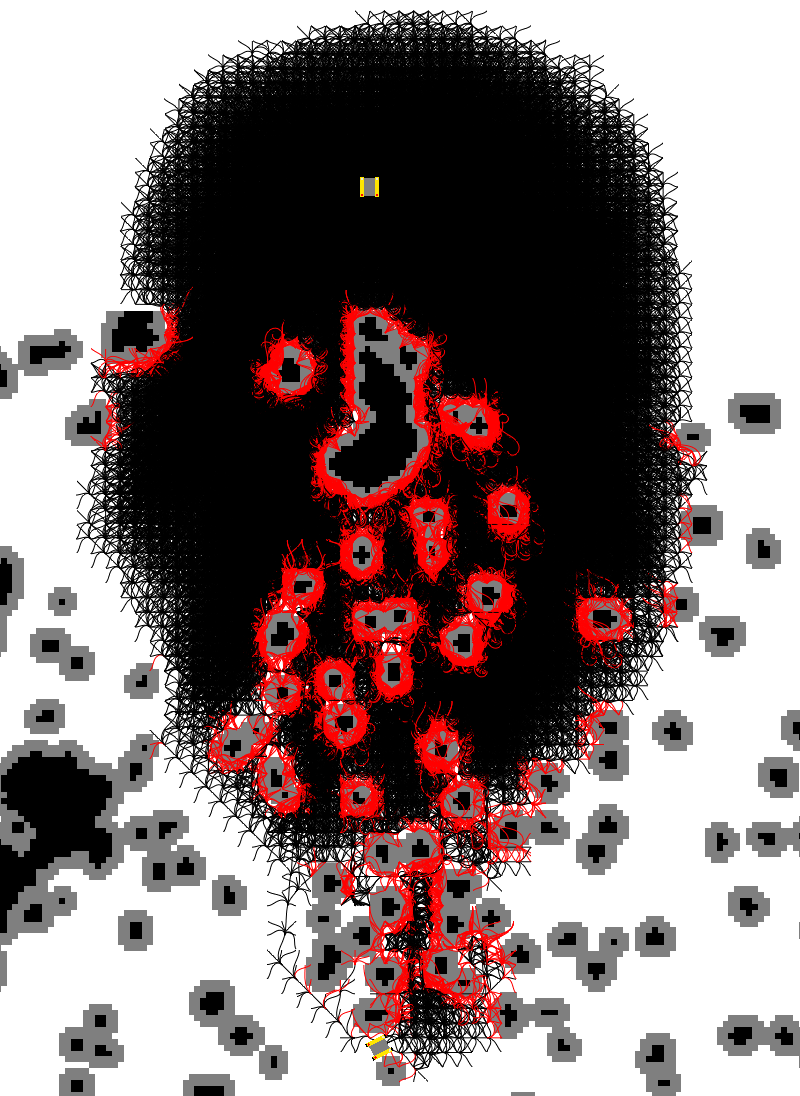}}
		\caption{}\label{fig:methodTwo}
	\end{subfigure}
	\begin{subfigure}{0.24\textwidth}
		\fbox{\includegraphics[height=\textwidth,angle=-90]{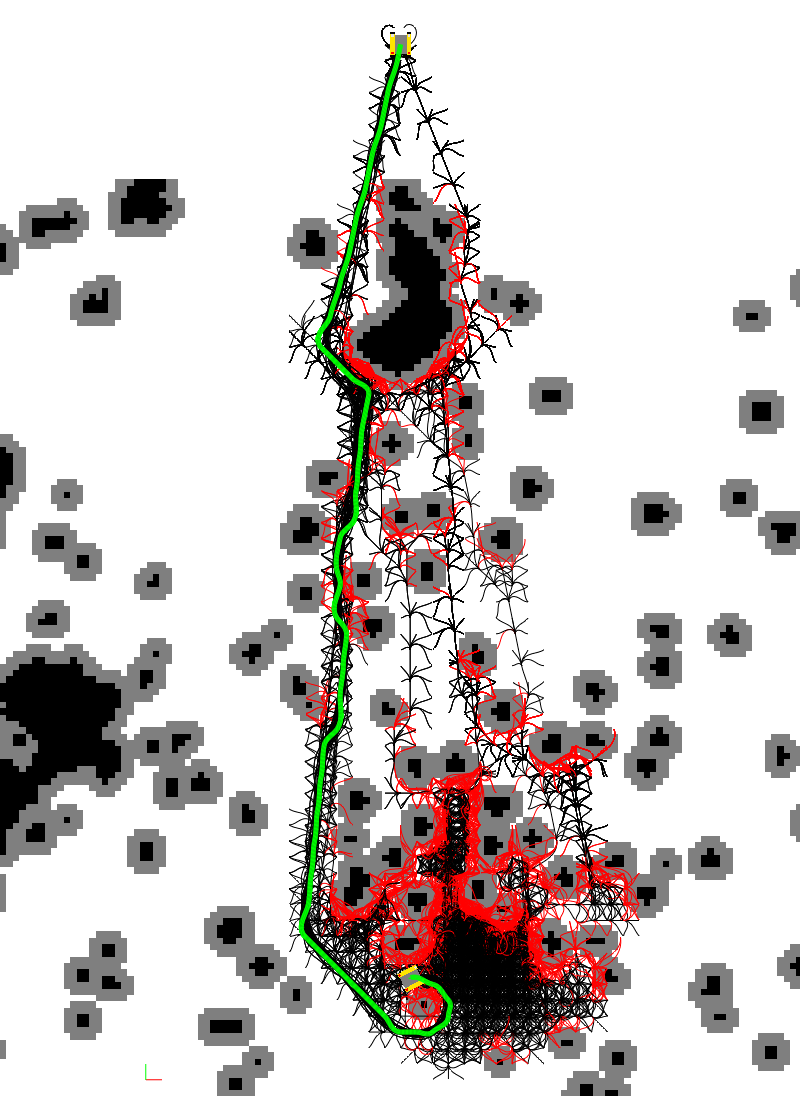}}
		\caption{}\label{fig:methodThree}
	\end{subfigure}
	\end{center}
	\vspace{-.5cm}
	\caption{\footnotesize{Search spaces for \textbf{(a)} the baseline with no rollouts (NR), \textbf{(b)} PER, \textbf{(c)} GER, \textbf{(d)} GEGR. The start and goal states are depicted as a yellow robot on the left and right of each image respectively. Valid and invalid expansions of KEASL are marked in black and red respectively. NR and GEGR were the only methods able to find a solution (green path) within the one-second planning time constraint.}}
	\label{fig:comparison}
\end{figure}

\vspace{-.3cm}
In this work, we present three methods for incorporating stochastic models of controller behavior into KEASL.
These approaches aim to account for real-world execution uncertainty by simulating noisy controller responses during search.
The baseline method, KEASL without stochastic modeling, is referred to as NR (No Rollout), and an example of its search space is shown in Figure \ref{fig:easlOne}.
The three proposed methods vary in how the controller rollouts are integrated into search.
The first, Per-Expansion Rollout (PER), simulates rollouts for every expansion.
The second, Goal-Edge Rollout (GER), implements a lazy search method by deferring rollouts until after an edge expands to the goal \cite{mandalika2019generalized}.
The final, Goal-Edge Graph Revision (GEGR), builds on GER by revising the graph after rollouts are performed.
Each method progressively reduces the computational burden introduced by the simulated rollouts during search.

In the context of this work, the controller rollouts are defined as the series of states derived from a pure-pursuit controller \cite{Coulter_1992_278} with a constant lookahead, following a trajectory from a start to a goal position.
To generate the rollout, zero-mean Gaussian noise is added to the velocity commands determined by the controller, and the state is updated by integrating the command at a discretized time interval of 0.1 seconds until the goal state is achieved.
Multiple rollouts are performed to create a stochastic distribution of potential robot behavior in response to the planned trajectory.
If any rollout is in collision with an obstacle, the node is marked invalid, and is unable to be expanded in the future.
If no rollouts collide with an obstacle, the cost of the edge is updated based on the average duration estimate of the simulated controller behaviors.
We define a maximum allowable run-time of one second so that planned motions reflect the most recent version of the world map as it is updated during traversal. 

\vspace{-.5cm}
\FloatBarrier

\subsection{Per-Expansion Rollout (PER)}
\vspace{-.1cm}
The first method applies rollouts during every node expansion in the graph.
If all rollouts are collision-free, the node is expanded and its cost is updated.
Practically, this method is very computationally expensive, and leads to expansions focused around the initial node, especially in cluttered environments.
Figure \ref{fig:methodOne} shows the resulting search space, where a solution is unable to be found when using Anytime Repairing A$^*$ (ARA$^*$) \cite{ARA} with a maximum runtime of one second.
Algorithm \ref{alg:PER} shows the steps for incorporating the rollout analysis into search.
\vspace{-.3cm}
\begin{algorithm}[htb]
\scriptsize
\caption{PER (Per-Expansion Rollout)}\label{alg:PER}
\begin{algorithmic}[1]
	\STATE $OPEN \gets$ {start node}
	\WHILE{$OPEN$ not empty}
		\STATE $n \gets$ OPEN.pop()
		\STATE $CLOSED \gets n$
		\FORALL{$n'$ in Expand($n$)}
			\IF{\textbf{every} rollout($n'$) is valid}
				\STATE UpdateCost($n$, $n'$)
				\IF{$n$ is goal \AND Cost($n) < OPEN.top()$}
					{\RETURN extractSolution($n$)}
				\ENDIF
				\STATE checkAndInsert($n'$,OPEN)
			\ENDIF
		\ENDFOR
	\ENDWHILE
\end{algorithmic}
\end{algorithm}

\vspace{-1.0cm}
\subsection{Goal-Edge Rollout (GER)}
\vspace{-.1cm}
To overcome the computational complexity of PER, GER defers the sampling until a node expands to the goal.
At this point, controller rollouts are performed along the potential solution.
If any rollouts are in collision with an obstacle, the final node is marked invalid and search continues.
If there are no collisions, the cost of the final node is updated as the average of the estimated rollout durations.
If the updated cost remains lower than the next-best candidate on the open list, the solution is returned.
If it does not, the node is reinserted into the open list and search continues.
Figure \ref{fig:methodTwo} shows the search space where a solution is unable to be found within one second because expansions are exhausted radially from the goal due to a rollout collision toward the beginning of search.
If the collision were to occur closer to the goal, a solution is more likely to be returned, but this is unable to be guaranteed in unstructured environments.
Algorithm \ref{alg:GER} shows the process for including the deferred rollouts in the search process.
\vspace{-.5cm}
\begin{algorithm}[hb]
	\scriptsize
	\caption{GER (Goal-Edge Rollout)}\label{alg:GER}
	\begin{algorithmic}[1]
		\STATE $OPEN \gets$ {start node}
		\WHILE{$OPEN$ not empty}
				\STATE $n \gets$ OPEN.pop()
				\STATE $CLOSED \gets n$
				\IF{$n$ is goal}
					\IF{\textbf{every} rollout(start node, $n$) is valid}
						\STATE Update goal-edge cost as the average of the rollouts
						\IF{Cost($n) < OPEN.top()$}
							{\RETURN extractSolution($n$)}
						\ENDIF
						\STATE Reinsert goal node into OPEN
					\ENDIF
				\ENDIF
				\FORALL{$n'$ in Expand($n$)}
					\STATE Cost($n$, $n'$)
					\STATE checkAndInsert($n'$,OPEN)
				\ENDFOR
		\ENDWHILE
	\end{algorithmic}
\end{algorithm}

\subsection{Goal-Edge Graph Revision (GEGR)}
The third iteration of this work uses the idea of GER, but instead of simply marking the candidate edge invalid, looks back through the graph and removes all descendants of the node corresponding to the point of collision from the rollouts.
The steps are shown in Algorithm \ref{alg:graphRevision}.
By removing nodes descending from the point of collision, search can focus on expansions in areas where all rollouts are valid.
Figure \ref{fig:methodThree} shows the resulting search space and solution (in green) from incorporating the graph revision step.
\vspace{-.5cm}
\begin{algorithm}
	\scriptsize
	\caption{Graph Revision}\label{alg:graphRevision}
	\begin{algorithmic}[1]
		\STATE $n_c$ = collision node
		\STATE $n_c$.RemoveNodeDescendents($OPEN$,$CLOSED$)
		\STATE $n_c$.MarkInvalid()
		\STATE $CLOSED \gets n_c$
	\end{algorithmic}
\end{algorithm}

\vspace{-.5cm}
The graph revision occurs after rollouts are performed, and if any rollout collides with an obstacle.
The full GEGR algorithm is shown in Algorithm \ref{alg:GEGR}, with the graph revision component shown in Algorithm \ref{alg:graphRevision}.
\vspace{-.5cm}
\begin{algorithm}[htb]
\scriptsize
\caption{GEGR (Goal-Edge Graph Revision)}\label{alg:GEGR}
\begin{algorithmic}[1]
	\STATE $OPEN \gets$ {start node}
	\WHILE{$OPEN$ not empty}
		\STATE $n \gets$ OPEN.pop()
		\STATE $CLOSED \gets n$
		\IF{$n$ is goal}
			\IF{\textbf{every} rollout(start node, $n$) is valid}
				\STATE Update goal-edge cost as the average of the rollouts
				\IF{Cost($n) < OPEN.top()$}
					{\RETURN extractSolution($n$)}
				\ENDIF
			\ELSE
				\STATE Perform the graph revision step 
				\STATE \textbf{continue}	
				\STATE Reinsert goal node into OPEN
			\ENDIF
		\ENDIF
		\FORALL{$n'$ in Expand($n$)}
			\STATE Cost($n$, $n'$)
			\STATE checkAndInsert($n'$,OPEN)
		\ENDFOR
	\ENDWHILE
\end{algorithmic}
\end{algorithm}
\vspace{-1.1cm}
\section{Experimental Design}\label{sec:experimentalDesign}
\vspace{-.2cm}
Six sets of experiments were performed to generate planning metrics from NR and GEGR.
Pure-pursuit control was used to perform rollouts instead of a more sophisticated method because the deterministic output and lack of obstacle awareness provide consistent behavior for the experiments.
Lookahead distance and obstacle inflation radius were used as the independent parameters for two experiments performed under three different mapping conditions (six experiments total).
The purpose of the lookahead experiments is to simulate a controller with varying levels of path-tracking ability.
The lookahead values range from 0.5 to 3.0 meters, in 0.5-meter increments.
In the obstacle expansion experiments, GEGR used a consistent lookahead of one meter, and the nominal environment map without obstacle inflation.
The obstacles in the map given to NR were inflated at radii spanning from 0.4 to 1.2 meters, in 0.4 meter increments to match the map resolution.
These expansion experiments analyze the performance of our methods compared to this straightforward approach of accounting for deviations in robot motions.
For every experiment, ARA$^*$ was used with a one-second planning time constraint, and if a solution was generated, 100 additional rollouts were performed to simulate a robot following the planned trajectory.
In the obstacle expansion experiments, these additional rollouts were performed on the nominal map for both NR and GEGR.
The rollouts in collision with an obstacle were tallied for each planning cycle.

\vspace{-0.5cm}
\begin{figure}
	\includegraphics[width=.19\textwidth]{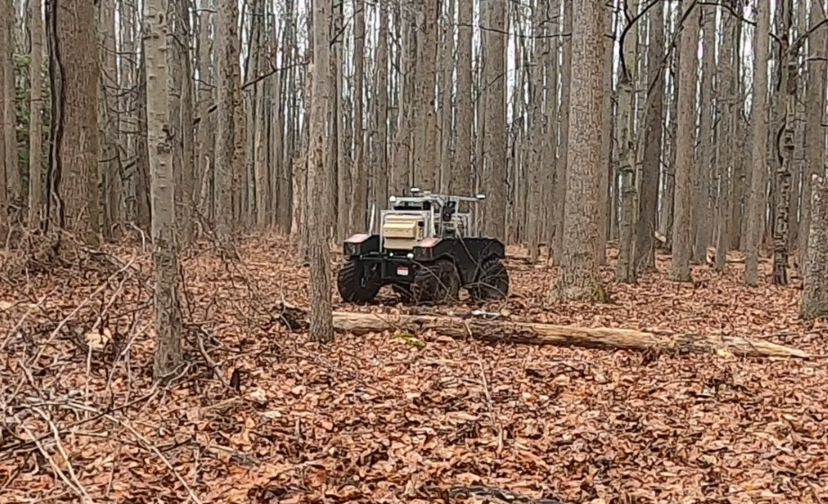}
	\includegraphics[width=.19\textwidth]{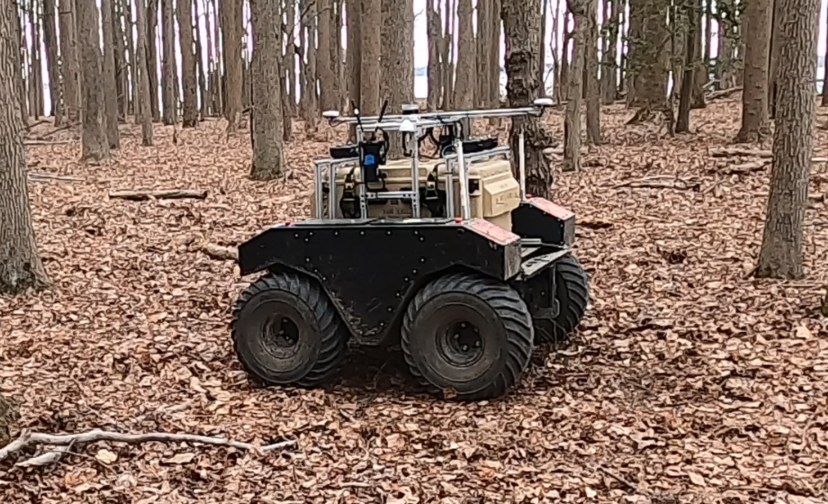}
	\includegraphics[width=.19\textwidth]{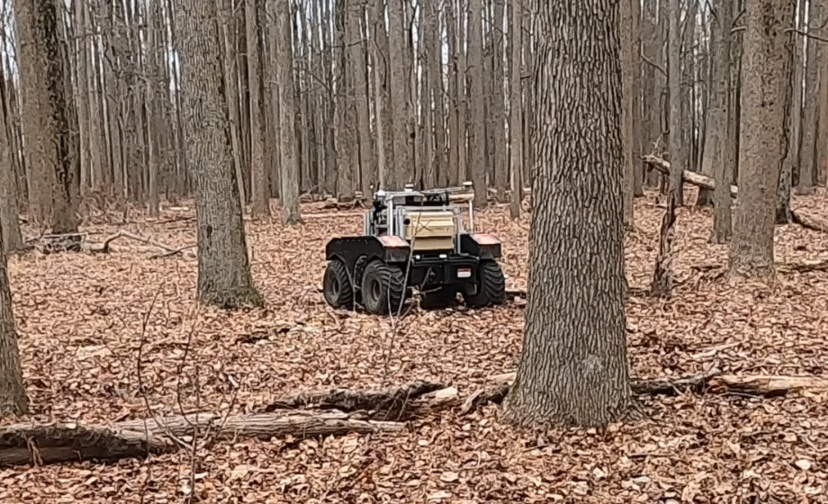}
	\includegraphics[width=.19\textwidth]{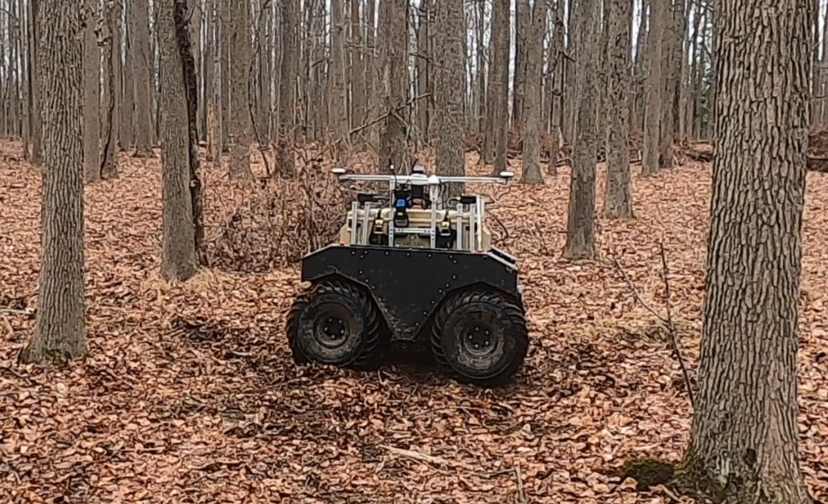}
	\includegraphics[width=.19\textwidth]{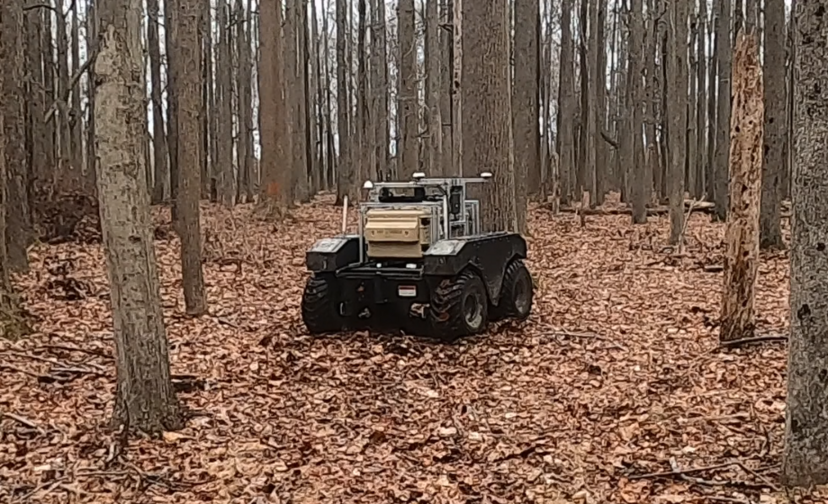}
	\vspace{-.2cm}
	\caption{\footnotesize{UGV in an unstructured, off-road environment during experiments.}}\label{fig:warthogWoods}
	\vspace{-.6cm}
\end{figure}

\vspace{-.5cm}
\subsection{Mapping Conditions}
\begin{figure}
	\centering
	\begin{subfigure}{.9\textwidth}
		\includegraphics[width=.19\textwidth]{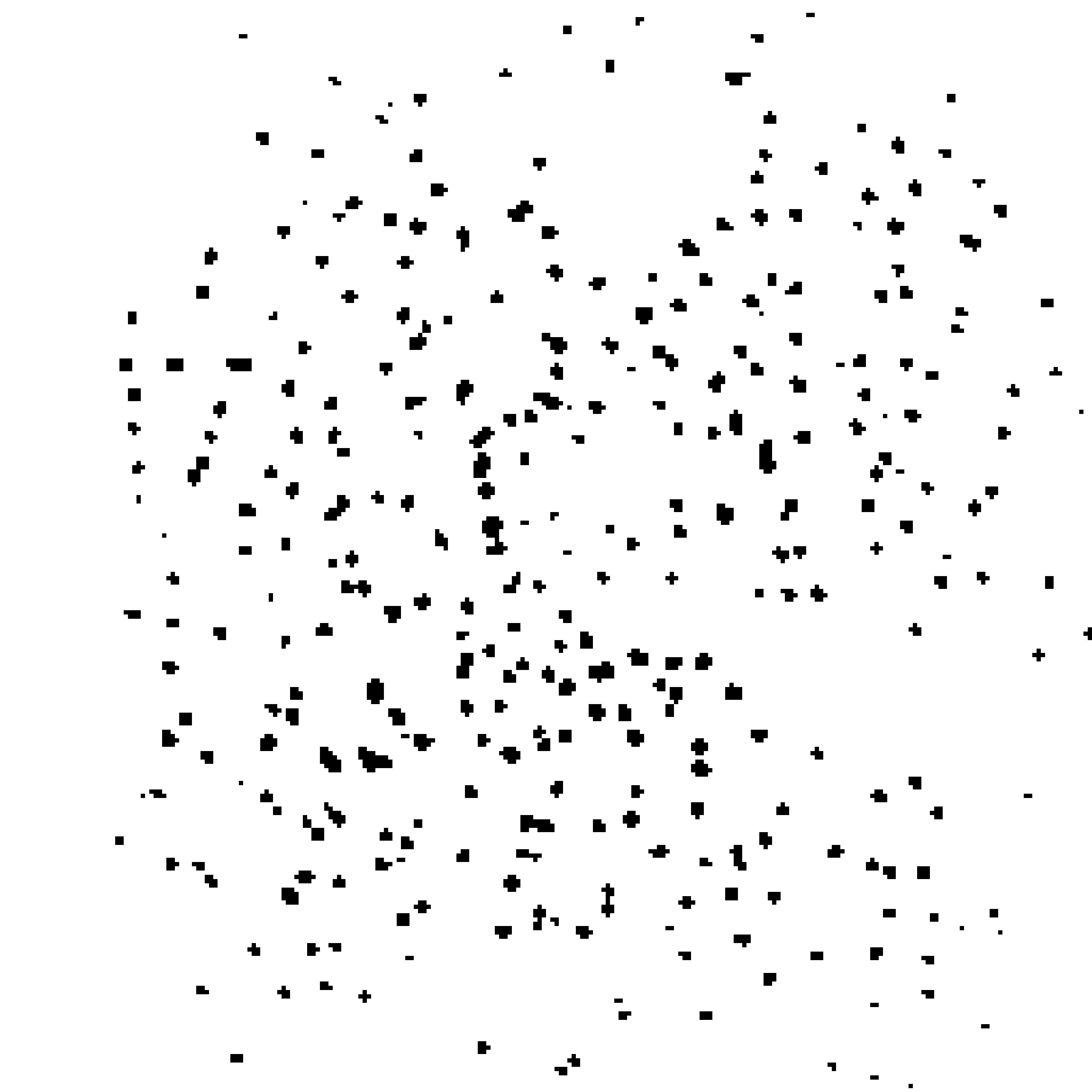}
		\includegraphics[width=.19\textwidth]{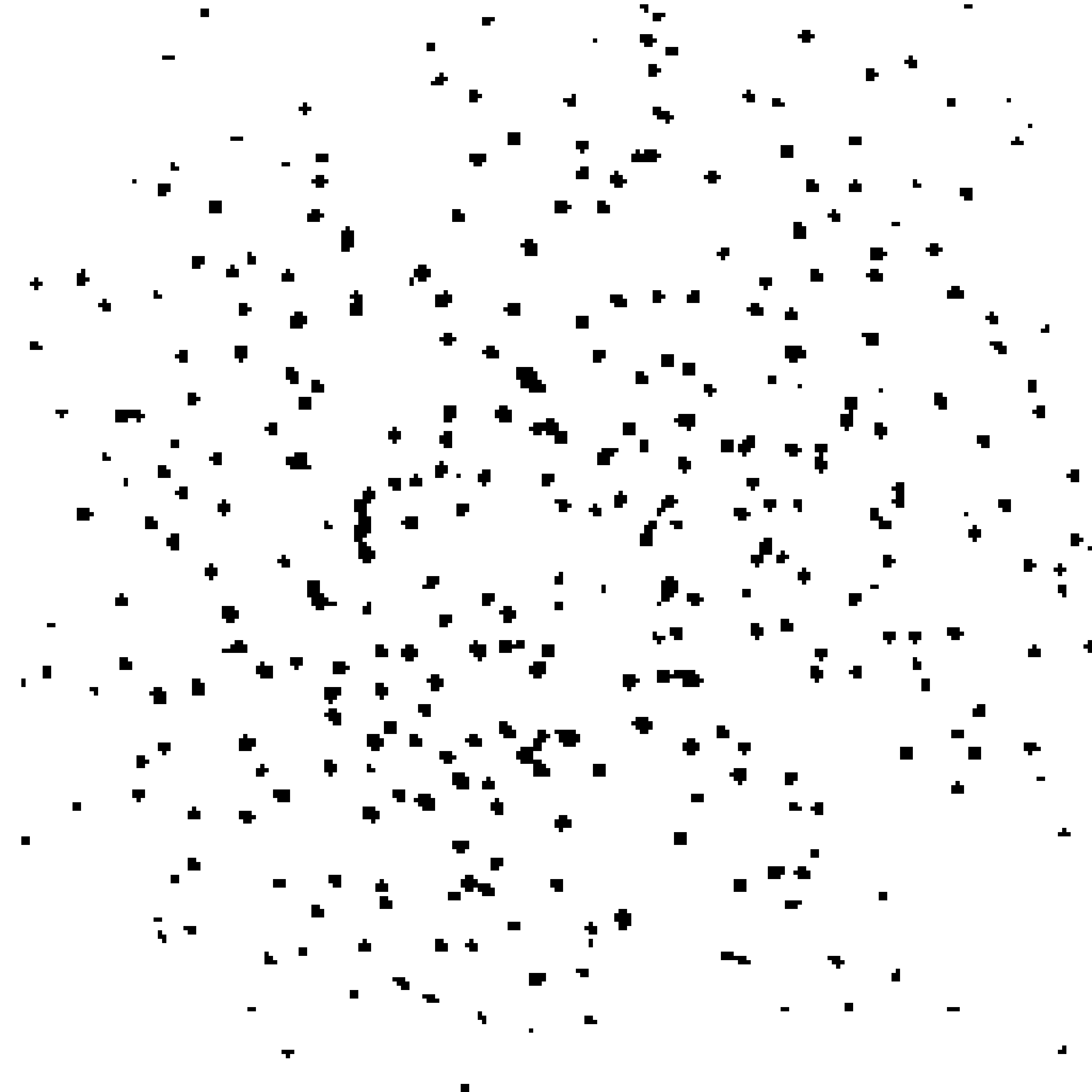}
		\includegraphics[width=.19\textwidth]{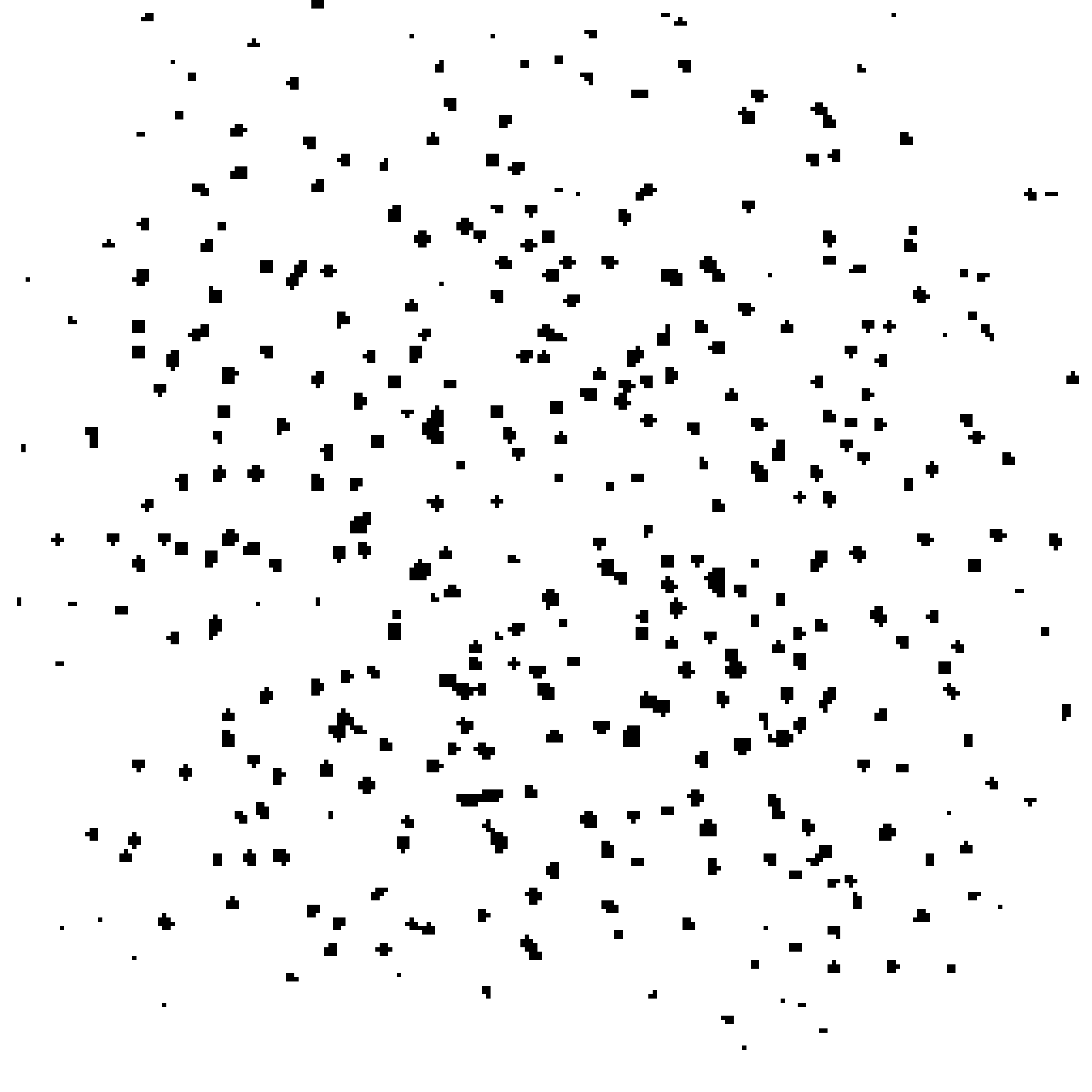}
		\includegraphics[width=.19\textwidth]{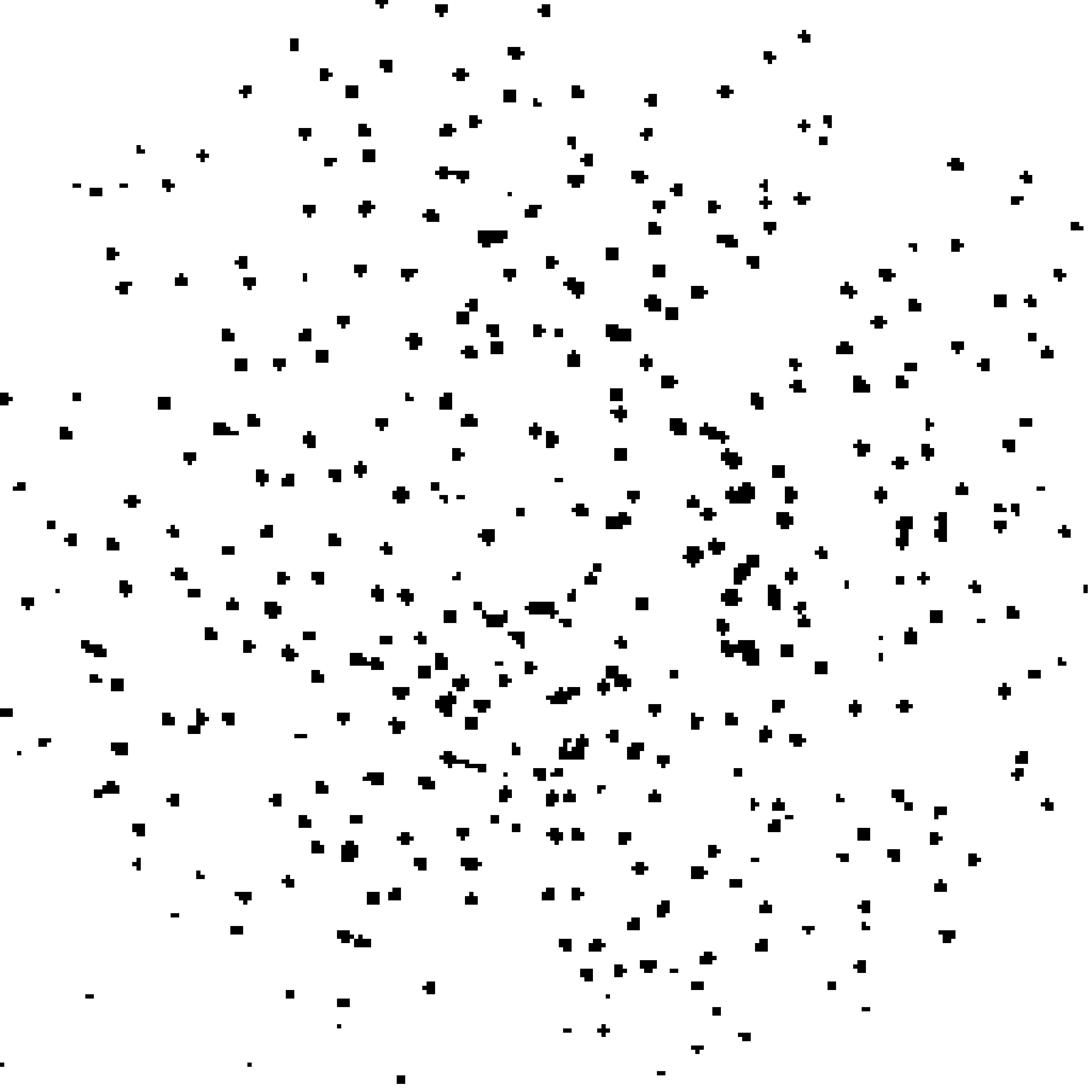}
		\includegraphics[width=.19\textwidth]{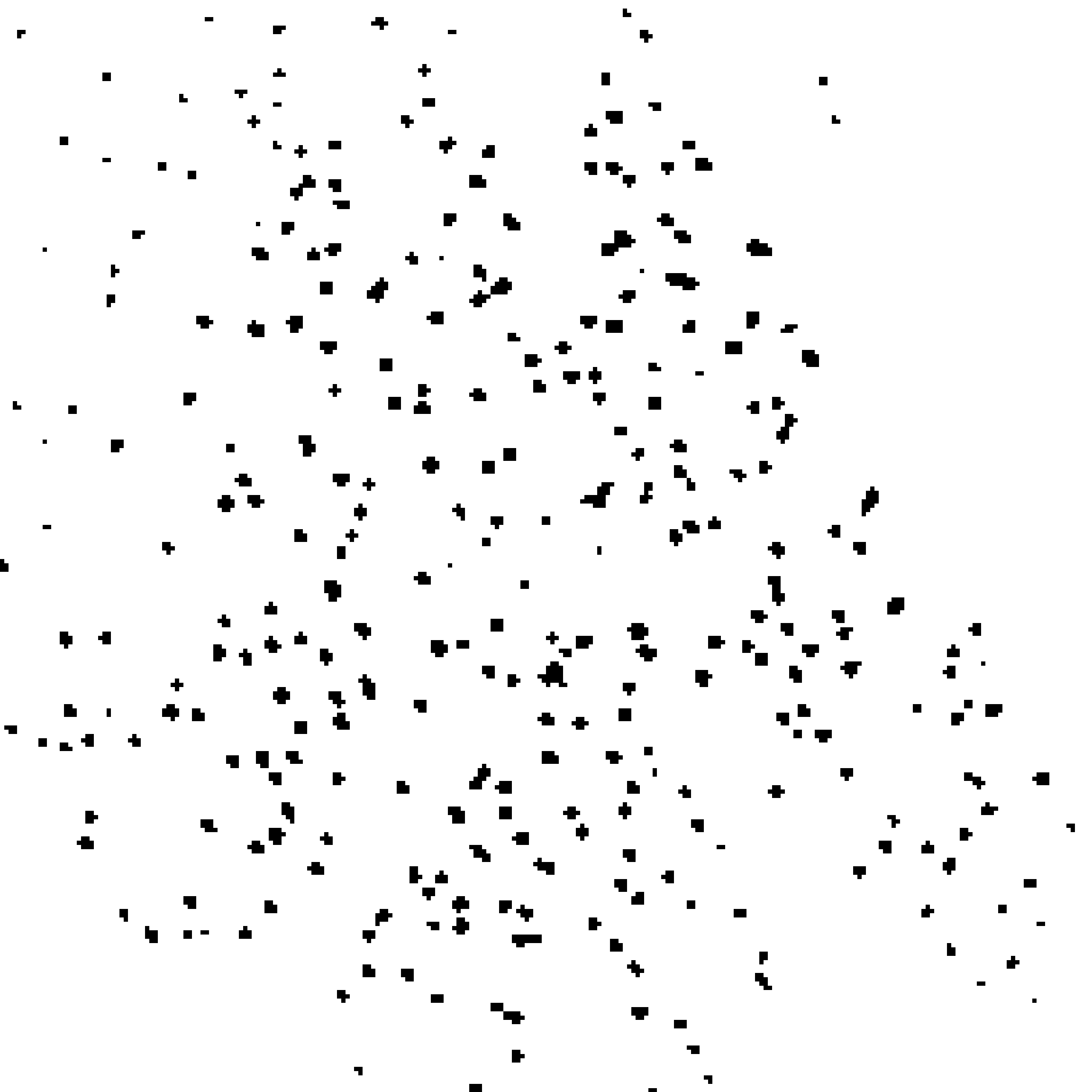}
	\end{subfigure}
	\begin{subfigure}{.9\textwidth}
		\includegraphics[width=.19\textwidth]{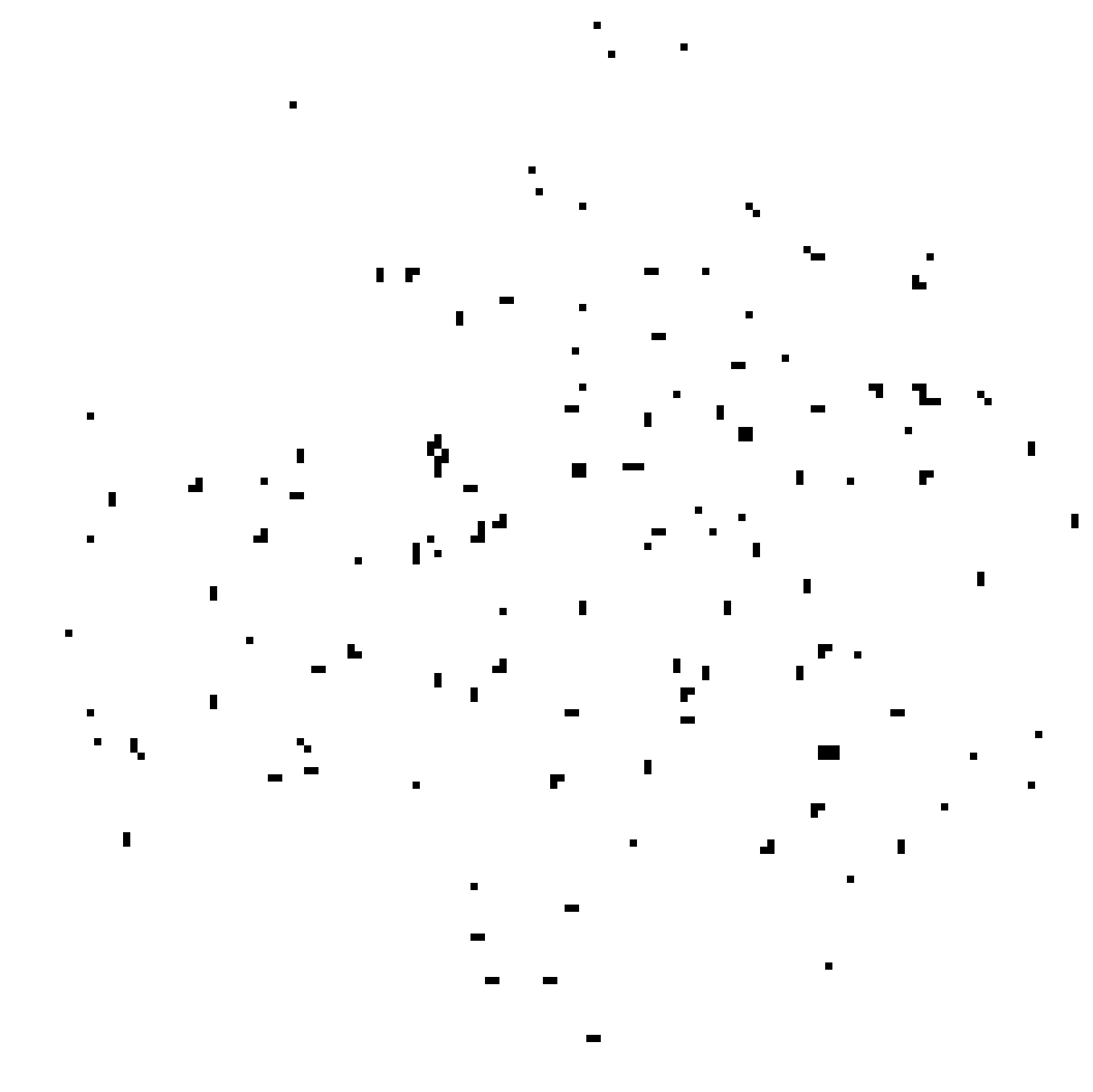}
		\includegraphics[width=.19\textwidth]{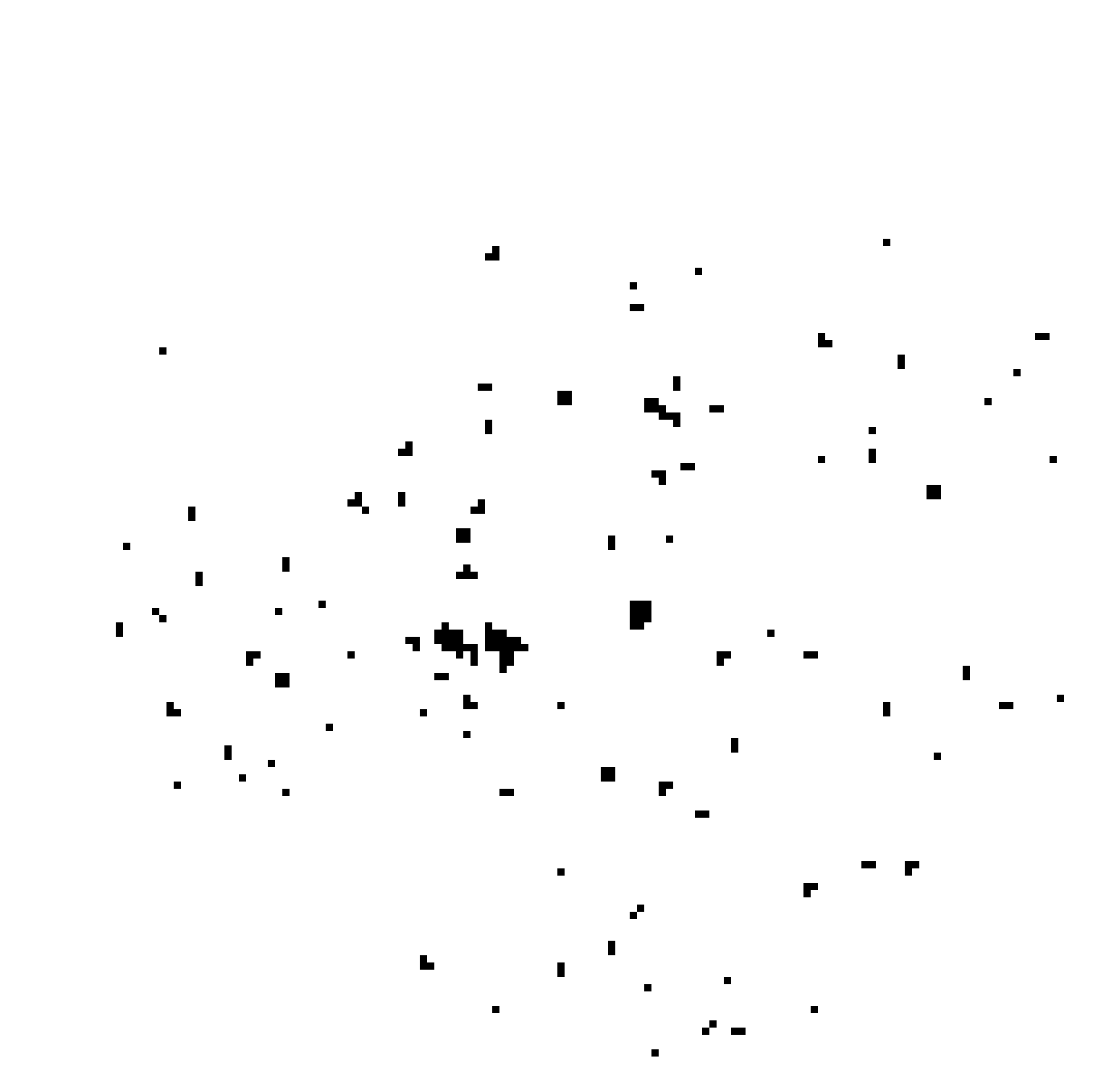}
		\includegraphics[width=.19\textwidth]{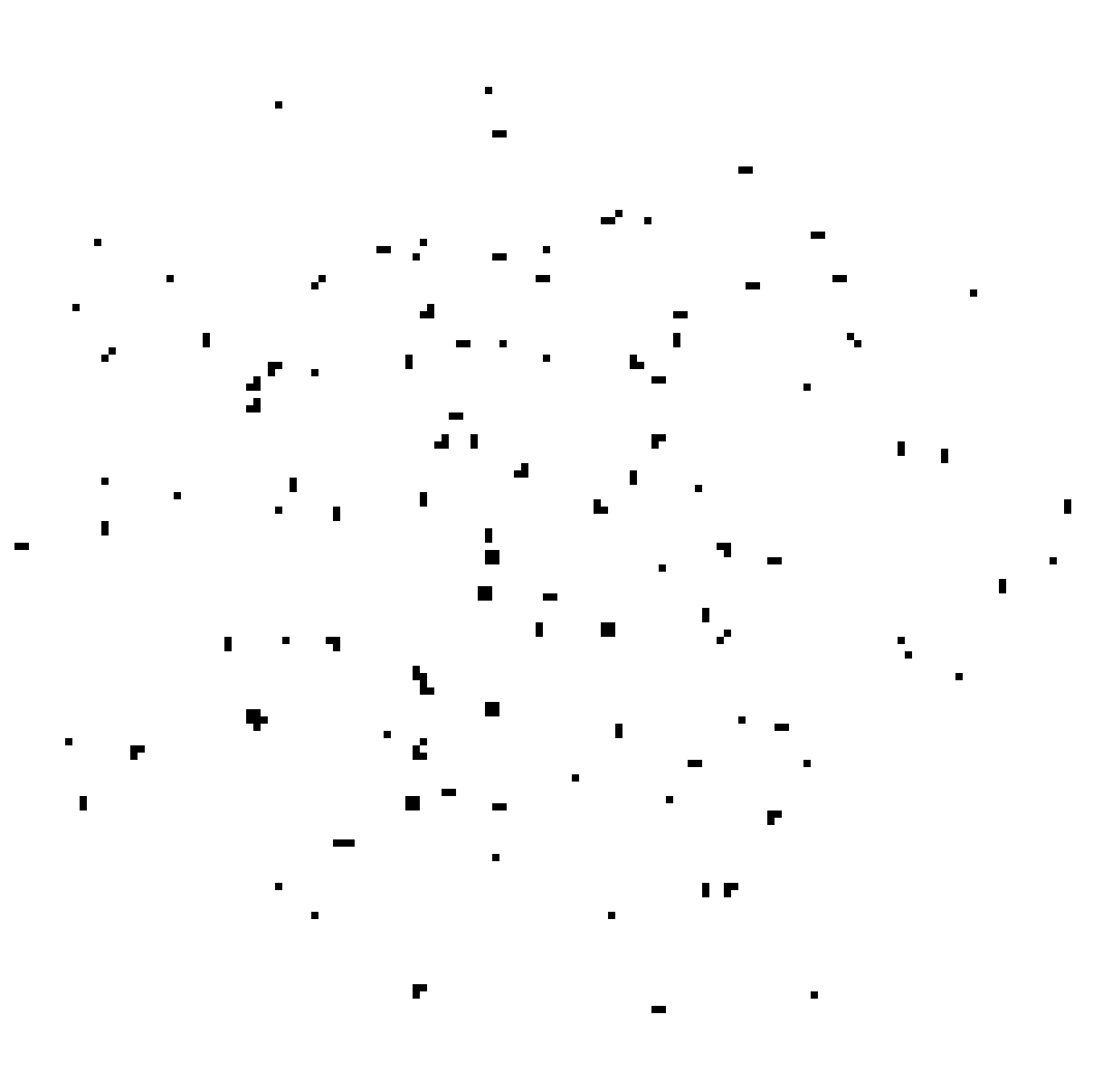}
		\includegraphics[width=.19\textwidth]{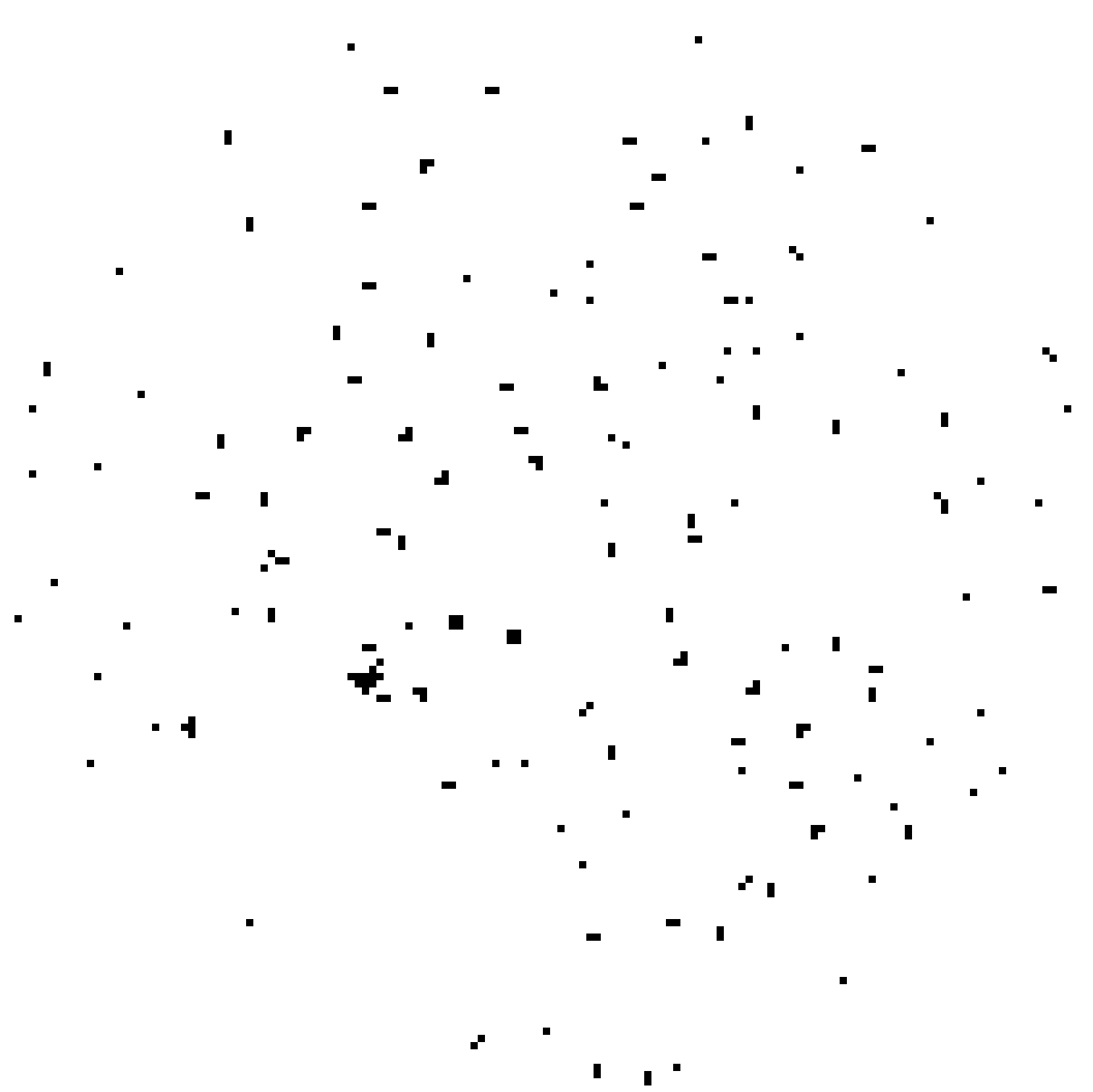}
		\includegraphics[width=.19\textwidth]{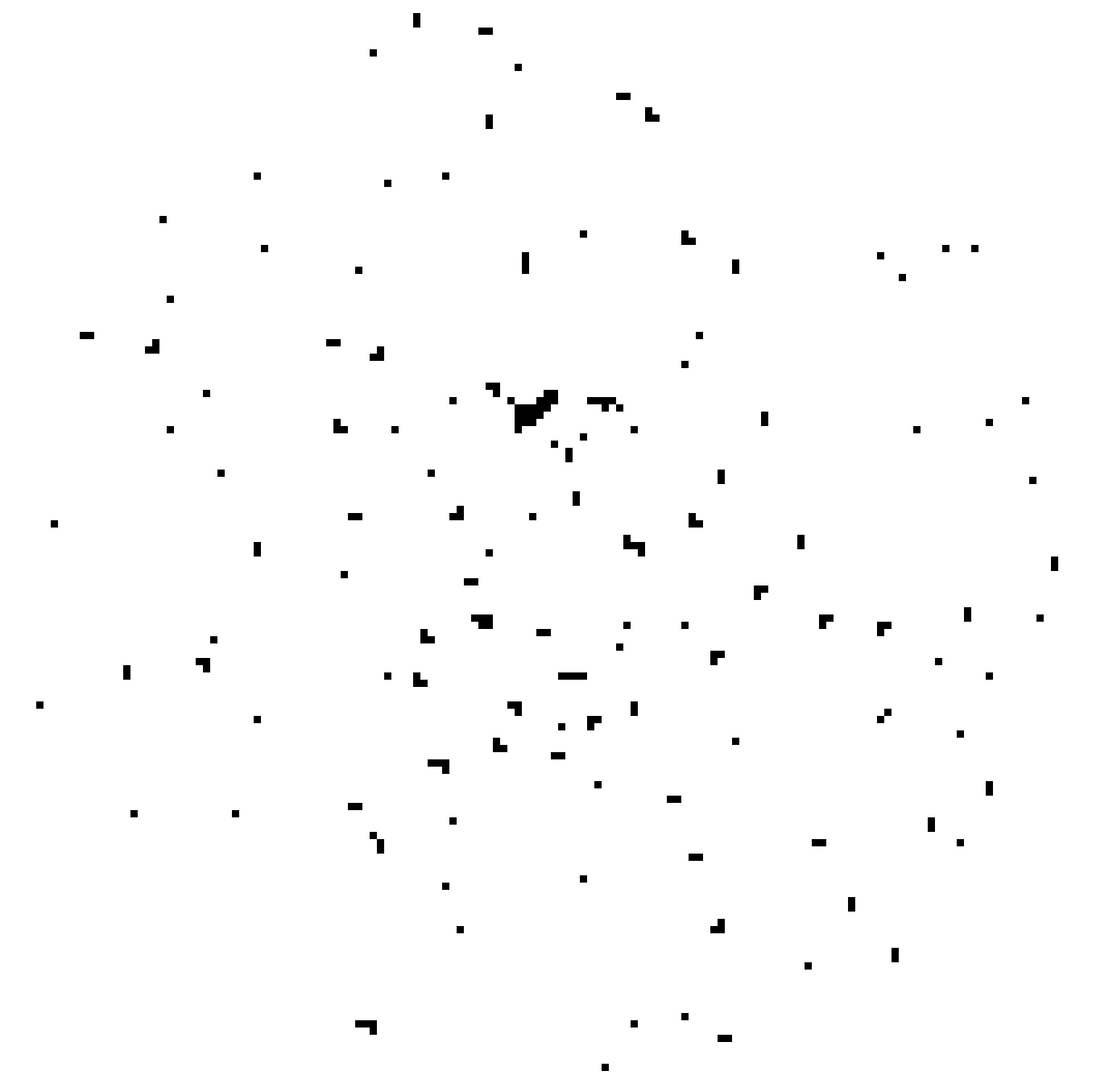}
	\end{subfigure}
	\caption{Top row: Map samples from Mapping Algorithm 1, each with a width and height of 102.4 meters. Bottom row: Map samples from Mapping Algorithm 2, each with a width and height of 60.4 meters.}
	\label{fig:mapSamples}
\end{figure}

Field experiments were performed on a UGV in an unstructured, offroad environment, shown in Figure \ref{fig:warthogWoods}.
From these experiments, two sets of planning problems (start state, goal state, environment map) were collected as the robot moved through the environment.
Each set was generated with a different mapping algorithm.
Mapping Algorithm 1 is a neural network approach (273 planning problems), and Mapping Algorithm 2 is a geometric approach (402 planning problems).
Samples of maps generated by each are shown in Figure \ref{fig:mapSamples}.

\vspace{-.3cm}
\vspace{-.5cm}

The outputs from each system are similar because they are both being used in the same environment, but they differ in their assessment of unobserved regions.
Mapping Algorithm 1 predicts the presence of obstacles in obscured areas because the neural network is able to fill in the gaps based on the context of the environment.
The output from Mapping Algorithm 2 is only able to fill in areas of the map that are directly observed.
Tests were performed using both mapping systems to show the consistency of our contributions under different perception configurations.

The ablation experiments serve to analyze the performance of GEGR with a full spectrum of environment map complexities.
45 maps were generated using Perlin noise \cite{perlin2002improving} for more realistic obstacle shapes and distributions, like in \cite{menonHAEASL}.
Each map contains a ring of obstacles with a 45-meter radius, and a 5-meter radius ``safe-zone'' around the starting point of the robot.
In each map, 8 goal locations were placed equally spaced 50 meters from the center starting point, creating 360 planning problems.

\begin{figure}
	\vspace{-.4cm}
	\includegraphics[width=.19\textwidth]{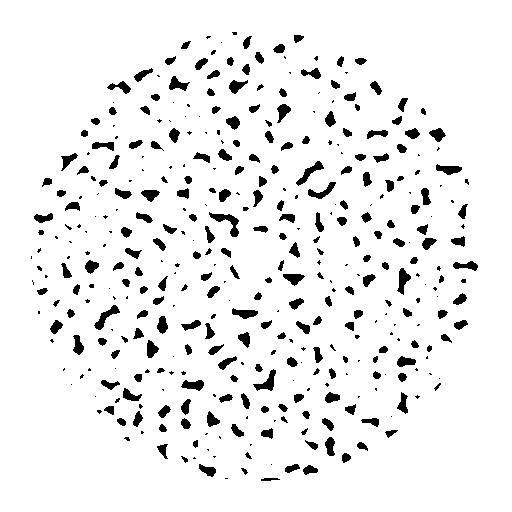}
	\includegraphics[width=.19\textwidth]{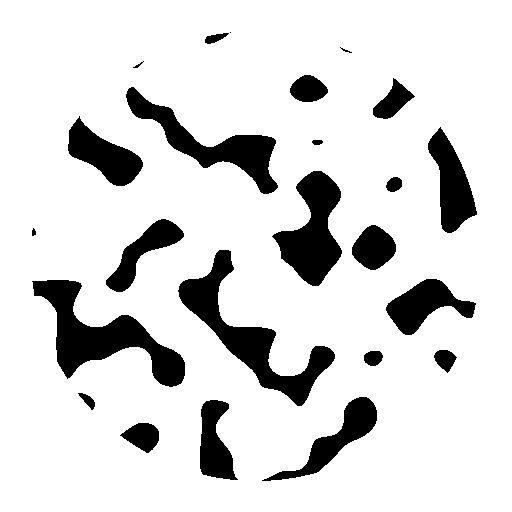}
	\includegraphics[width=.19\textwidth]{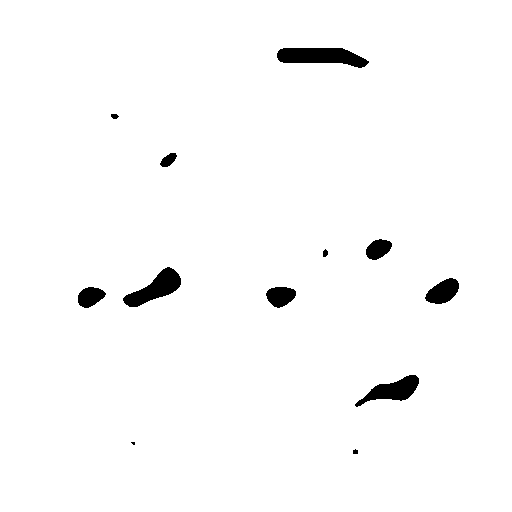}
	\includegraphics[width=.19\textwidth]{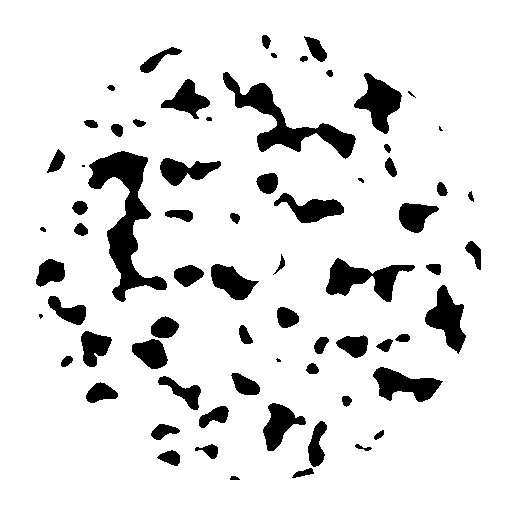}
	\includegraphics[width=.19\textwidth]{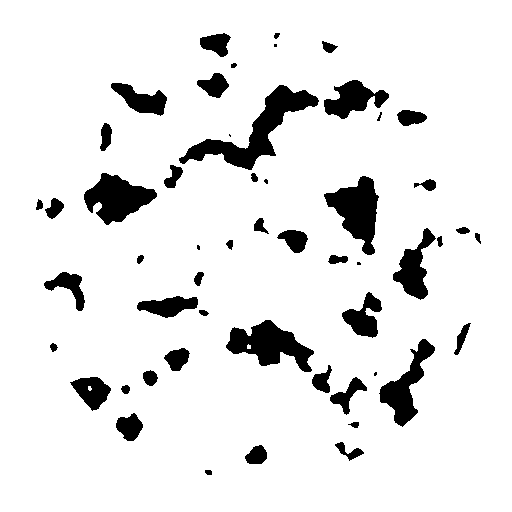}
	\caption{A sample of 5 of the 45 simulated maps used for the ablation experiments}\label{fig:perlinMaps}
\end{figure}

Figure \ref{fig:perlinMaps} depicts a sample of the 45 maps used for the experiments.
The wide variety of obstacle shapes, sizes, and distributions allows for an in-depth analysis of algorithm performance.
\section{Results}\label{sec:results}
The results of the obstacle expansion and lookahead experiments are shown below in Figures \ref{fig:lookaheadData} and \ref{fig:cSpaceData} respectively.
Each set of figures shows the results for the test sets with Mapping Algorithm 1, Mapping Algorithm 2, and the Perlin maps.

Initial field experiments with PER showed consistent failure to find solutions within a one-second planning time.
Initially, 30 controller rollouts were performed for each edge expansion, and no solutions were found.
The number of rollouts per edge was reduced to 5, and PER was able to find 2 solutions within the same time constraint.
The high computational cost of PER necessitated the development and shift to GER.
Although more successful in finding solutions, there was still a consistent inability to plan motions in cluttered environments.
Using 30 rollouts per goal-edge revision, GER generated 3 solutions.
Dropping the number of rollouts to 5 led to an increase in the number of solutions to 21 - still far less than a sufficient success rate.
The further development of GEGR was far more successful and succeeded in all cases using 30 rollouts per goal edge revision. 
This shift proved the importance of revising the graph in addition to revising goal edges within the graph.
The planning metrics discussed in this section omit the results of PER and GER because of the large gap in performance when compared to NR and GEGR.

\begin{figure}[H]
	\centering
	\includegraphics[width=\textwidth]{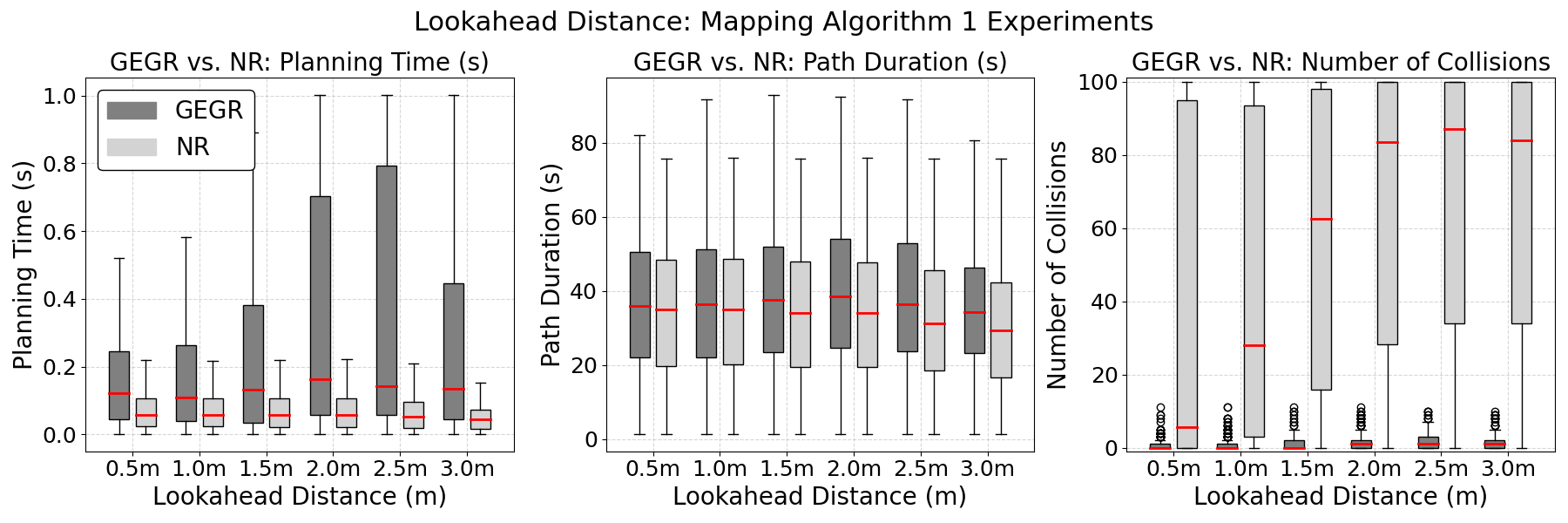}
	\includegraphics[width=\textwidth]{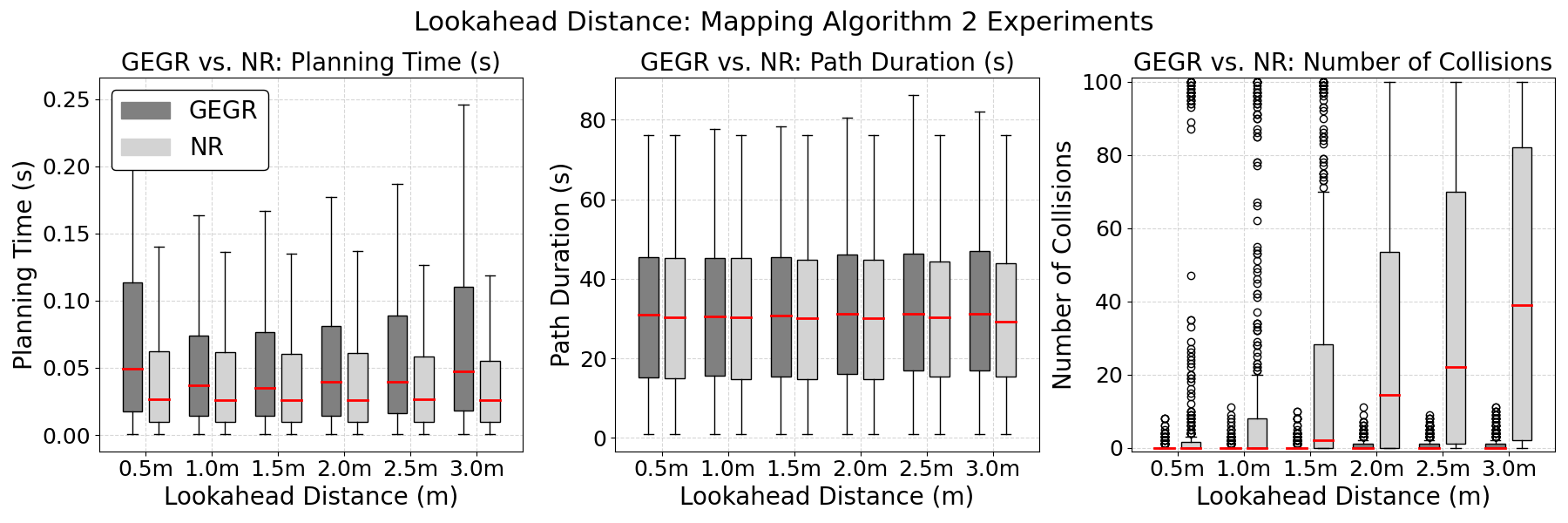}
	\includegraphics[width=\textwidth]{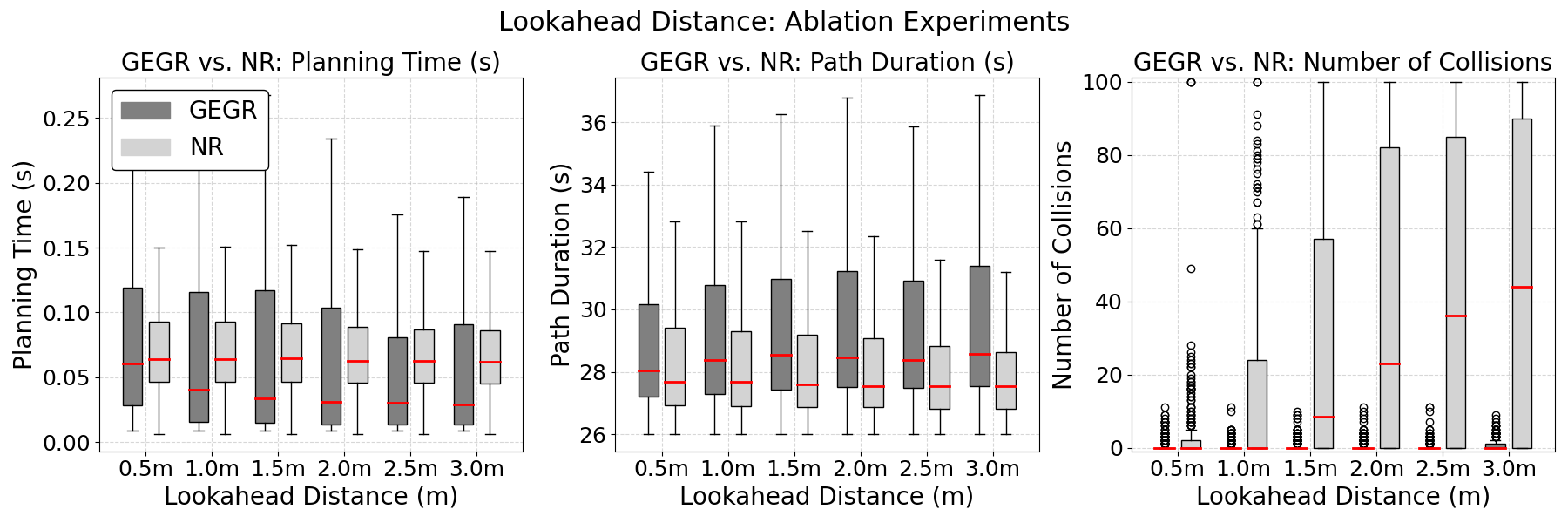}
	{\scriptsize
	\begin{tabular}{
		>{\centering\arraybackslash}p{1.0cm} || >{\centering\arraybackslash}p{1.4cm} | >{\centering\arraybackslash}p{1.4cm} || >{\centering\arraybackslash}p{1.4cm} | >{\centering\arraybackslash}p{1.4cm} || >{\centering\arraybackslash}p{1.4cm} | >{\centering\arraybackslash}p{1.4cm} ||}
		\multicolumn{7}{c}{}\\ 
		\hline
		\multicolumn{7}{c}{Number of Failed Planning Attempts}\\ 
		\hline
		LD (m) & \multicolumn{2}{c||}{Mapping Algorithm 1} & \multicolumn{2}{c||}{Mapping Algorithm 2} & \multicolumn{2}{c||}{Ablation} \\
		& GEGR & NR & GEGR & NR & GEGR & NR\\
		\hline\hline
		0.5 & 6 & 0 & 6 & 0 & 6 & 0 \\
		1.0 & 7 & 0 & 11 & 0 & 11 & 0 \\
		1.5 & 17 & 0 & 16 & 0 & 21 & 0 \\
		2.0 & 39 & 0 & 23 & 0 & 30 & 0 \\
		2.5 & 72 & 0 & 35 & 0 & 42 & 0 \\
		3.0 & 125 & 0 & 59 & 0 & 48 & 0 \\
	\end{tabular}
	}
	\caption{\footnotesize{Planning Metrics of NR and GEGR using different lookahead values. \textbf{LD:} Lookahead distance in meters, \textbf{Gray boxes:} 25th to 75th percentile, \textbf{Red lines:} Median value (50th percentile), \textbf{Whiskers:} The range of data (excluding outliers), \textbf{Circles:} Outliers beyond 1.5 times the interquartile range (IQR) - included in collision plots for clarity. \textit{Note: Plots correspond to the planning problems where both GEGR and NR found solutions. Table metrics indicate planning cycles where one failed, and the other did not.}}}
	\label{fig:lookaheadData}
\end{figure}

\begin{figure}[H]
	\centering
	\includegraphics[width=\textwidth]{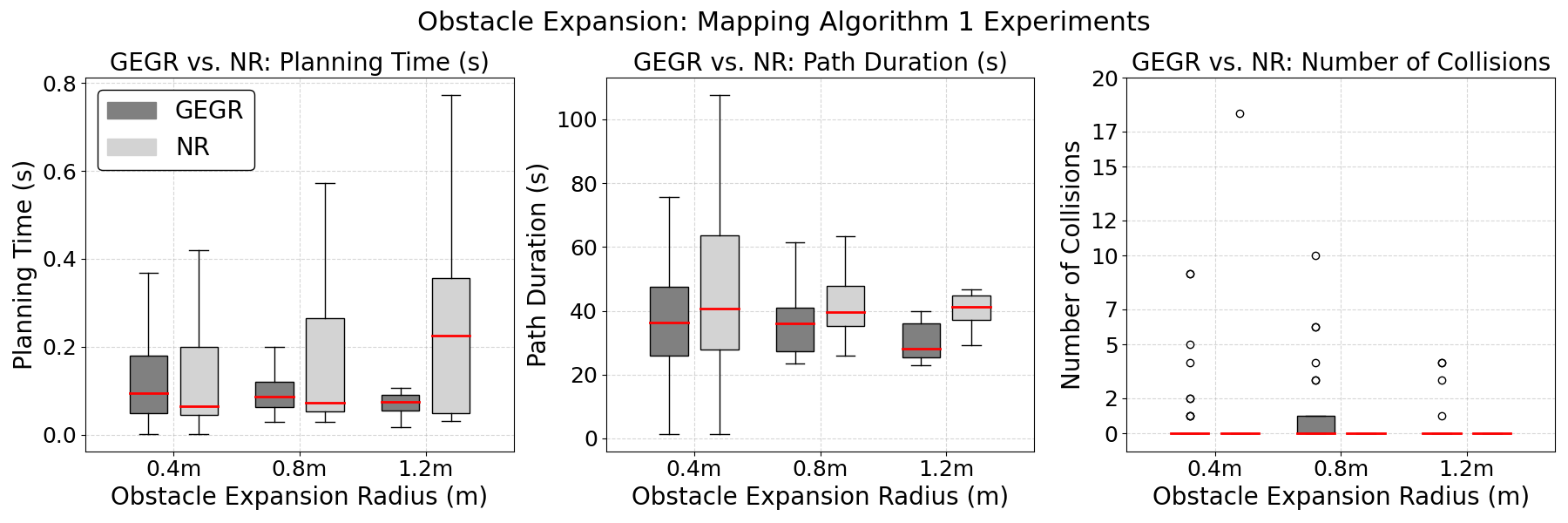}
	\includegraphics[width=\textwidth]{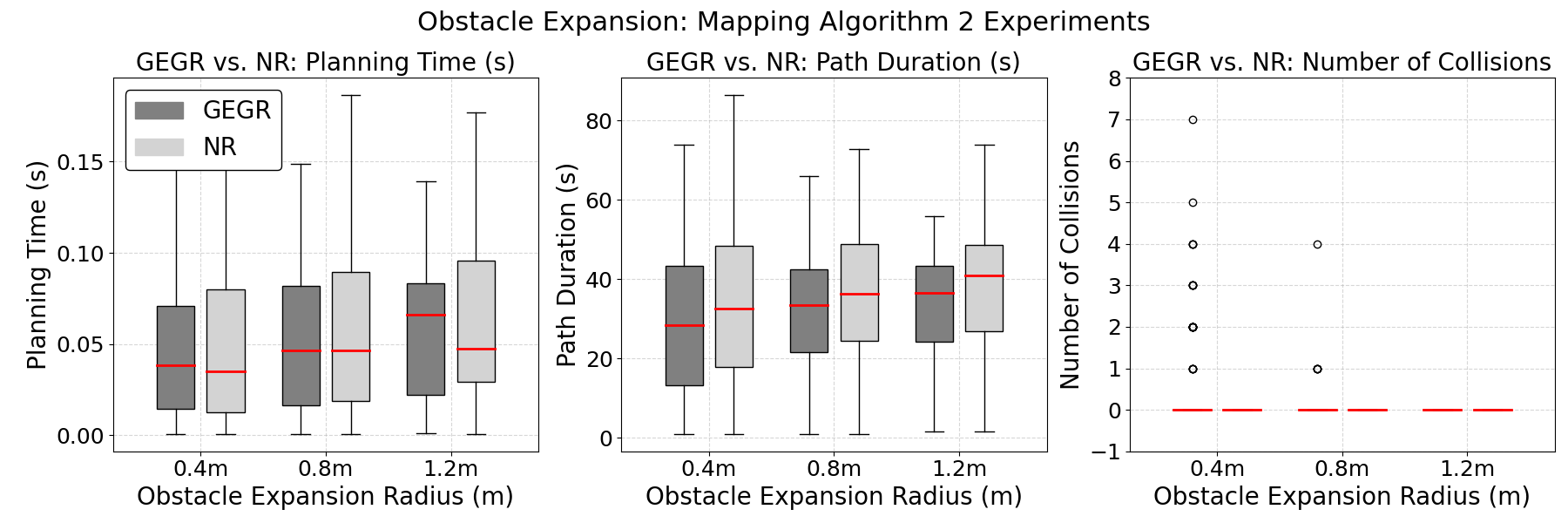}
	\includegraphics[width=\textwidth]{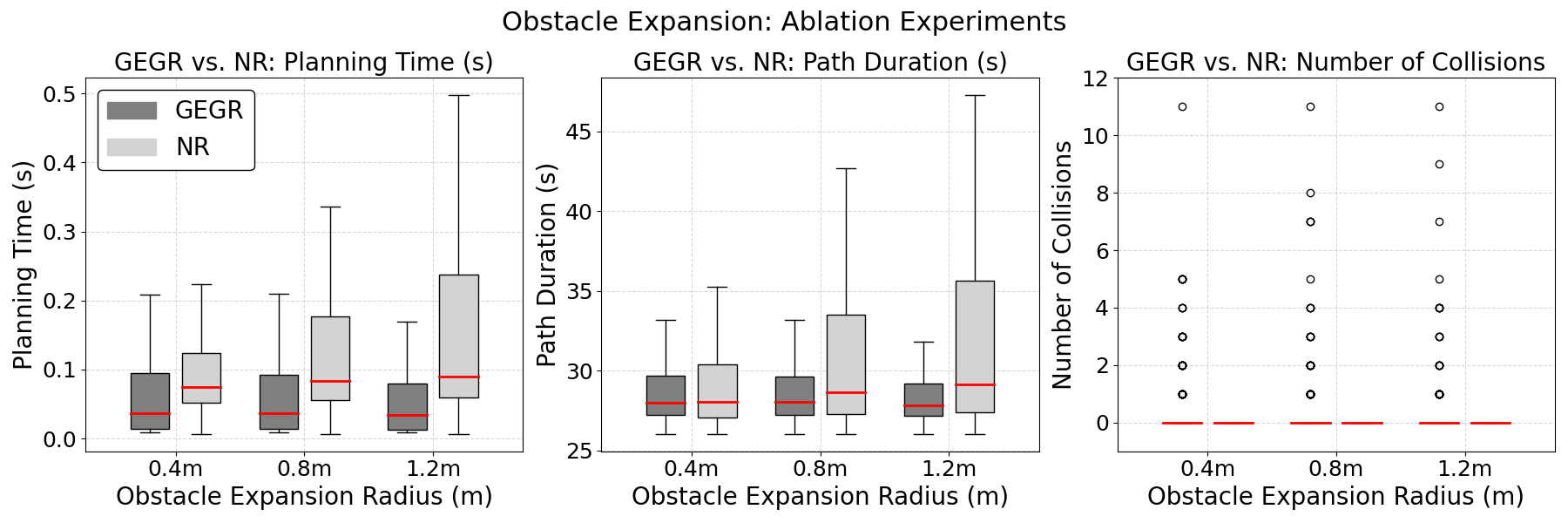}
	{\scriptsize
	\begin{tabular}{
		>{\centering\arraybackslash}p{1.0cm} || >{\centering\arraybackslash}p{1.4cm} | >{\centering\arraybackslash}p{1.4cm} || >{\centering\arraybackslash}p{1.4cm} | >{\centering\arraybackslash}p{1.4cm} || >{\centering\arraybackslash}p{1.4cm} | >{\centering\arraybackslash}p{1.4cm} ||}
		\multicolumn{7}{c}{}\\ 
		\hline
		\multicolumn{7}{c}{Number of Failed Planning Attempts}\\ 
		\hline
		OE (m) & \multicolumn{2}{c||}{Mapping Algorithm 1} & \multicolumn{2}{c||}{Mapping Algorithm 2} & \multicolumn{2}{c||}{Ablation} \\
		& GEGR & NR & GEGR & NR & GEGR & NR\\
		\hline\hline
		0.4 & 0 & 182 & 1 & 211 & 0 & 45 \\
		0.8 & 0 & 212 & 1 & 307 & 0 & 45 \\
		1.2 & 0 & 222 & 1 & 354 & 0 & 76 \\
	\end{tabular}
	}
	\caption{\footnotesize{Planning metrics of NR and GEGR using different levels of obstacle expansion. \textbf{OE:} Obstacle expansion radius in meters, \textbf{Gray boxes:} 25th to 75th percentile, \textbf{Red lines:} Median value (50th percentile), \textbf{Whiskers:} The range of data (excluding outliers), \textbf{Circles:} Outliers beyond 1.5 times the interquartile range (IQR) - included in collision plots for clarity. \textit{Note: Plots correspond to the planning problems where both GEGR and NR found solutions. Table metrics indicate planning cycles where one failed, and the other did not.}}}
	\label{fig:cSpaceData}
\end{figure}

\newpage
\subsection{Lookahead Experiments}
\vspace{-.6cm}
In pure-pursuit control, the lookahead distance acts as a gain for the level at which a given trajectory is followed.
These experiments use varying lookahead distances to simulate a controller following a path closely (with a 0.5m lookahead), to one which follows very loosely (with a 3.0m lookahead).
The results of these experiments are shown in Figure \ref{fig:lookaheadData}.
In all three sets, the number of collisions was higher for NR than GEGR, and increased with lookahead distance.
This is an expected result because a higher lookahead leads to more deviation from the planned trajectory.

\vspace{-0.7cm}
\begin{figure}[h]
	\centering
	\includegraphics[height=.47\textwidth,angle=-90,trim={10cm 0 10cm 0},clip]{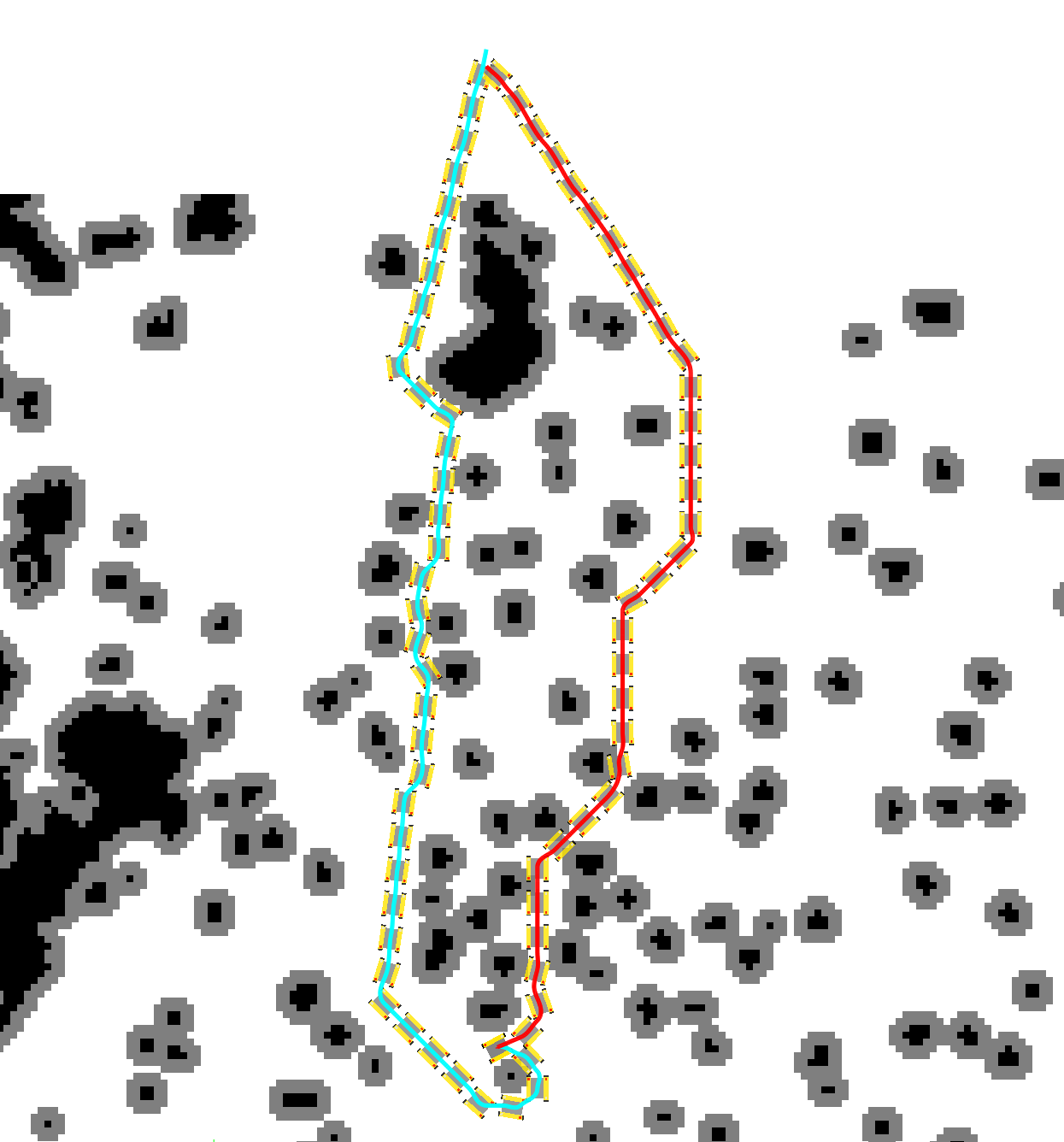}
	\includegraphics[height=.47\textwidth,angle=-90,trim={10cm 0 10cm 0},clip]{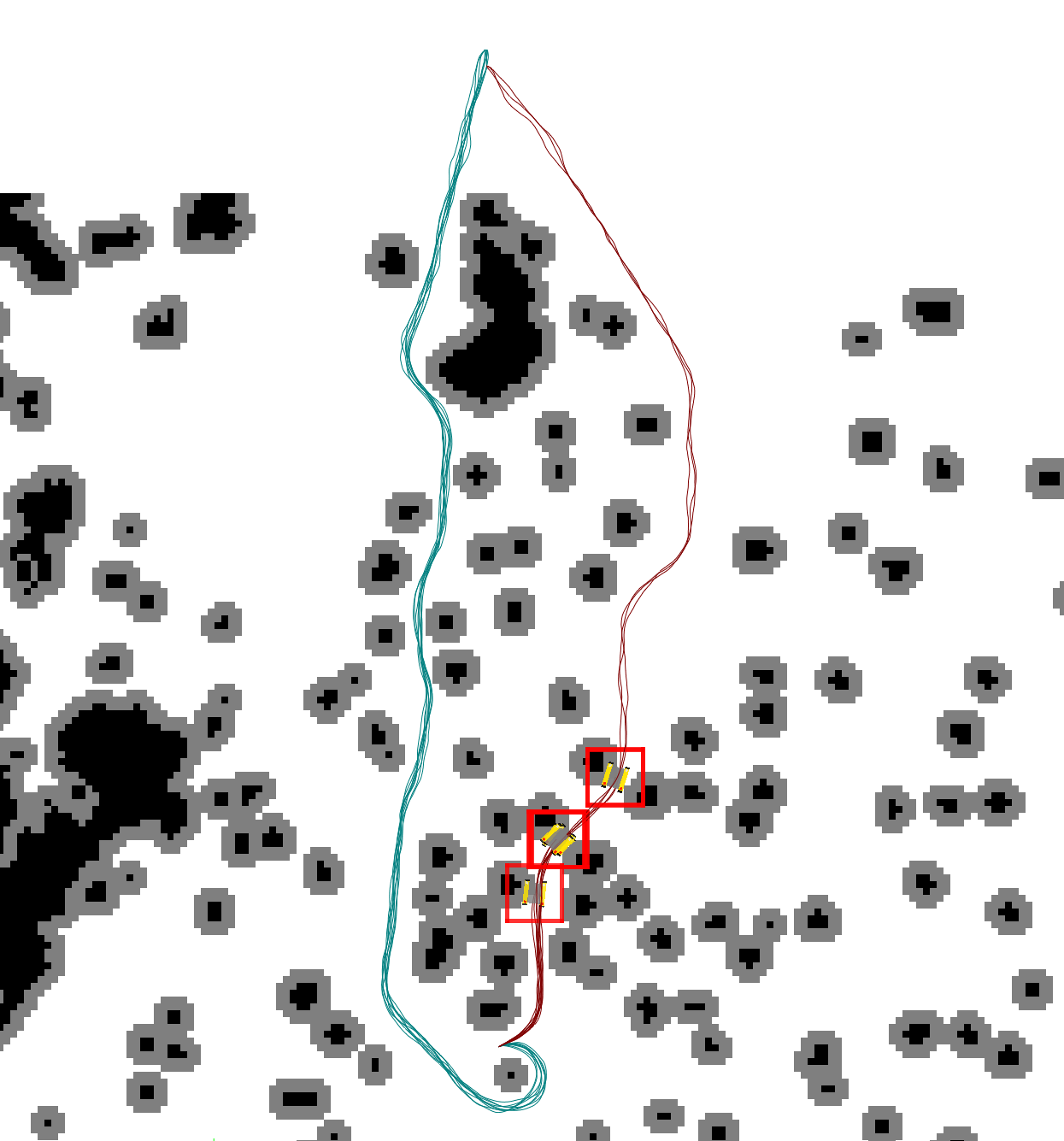}
	\vspace{-.2cm}
	\caption{\footnotesize{Left: Planned NR (red) and GEGR (cyan) trajectories with robot depictions. Right: 10 simulated controller rollouts on the solutions generated by NR (dark red) and GEGR (dark cyan) with collisions shown as robots surrounded by a red square. The map was generated by Mapping Algorithm 1.}}\label{fig:twoExamples}
\end{figure}
\vspace{-0.7cm}
The planning times of GEGR were higher than NR for every case because of the added computing overhead to perform the rollouts and revise the graph.
Additionally, the average path durations were higher for GEGR than NR, which is expected because NR finds more optimistic trajectories that may cut tightly between obstacles, and GEGR avoids those scenarios to account for possible controller deviations.
The left half of Figure \ref{fig:twoExamples} is a clear example of this trade-off, where the path by GEGR takes a less direct route through a more open area of the environment and NR chooses a more direct route between a dense clustering of obstacles.
The right half of the figure shows multiple simulated controller collisions in the segment of the NR trajectory that goes through the obstacle cluster.
Although the route generated by GEGR had a higher duration than NR, it considered potential path deviation by the controller and generated a plan with less likelihood for collision.
For both algorithms, the average times are well below the 1-second run-time and the path duration differences are marginal, indicating validity for field use.

The number of failed planning cycles by GEGR increased with lookahead distance due to the increasing difficulty of planning with rollouts from a controller with increasingly limited path following capability.
This illustrates the importance of predicting possible controller behavior in the motion planning process because a failure to plan under certain controller configurations indicates the need to evaluate the adequacy of the underlying control scheme.
The large increase in failed planning cycles for the 3.0-meter experiment with Mapping Algorithm 1 explains the drop in planning times, because the only successful solutions were likely with the most simple set of maps.

\vspace{-.3cm}
\subsection{Obstacle Expansion Experiments}
\begin{figure}[h]
	\centering
	\includegraphics[width=.2\linewidth,trim={0cm 0cm 0cm 2cm},clip]{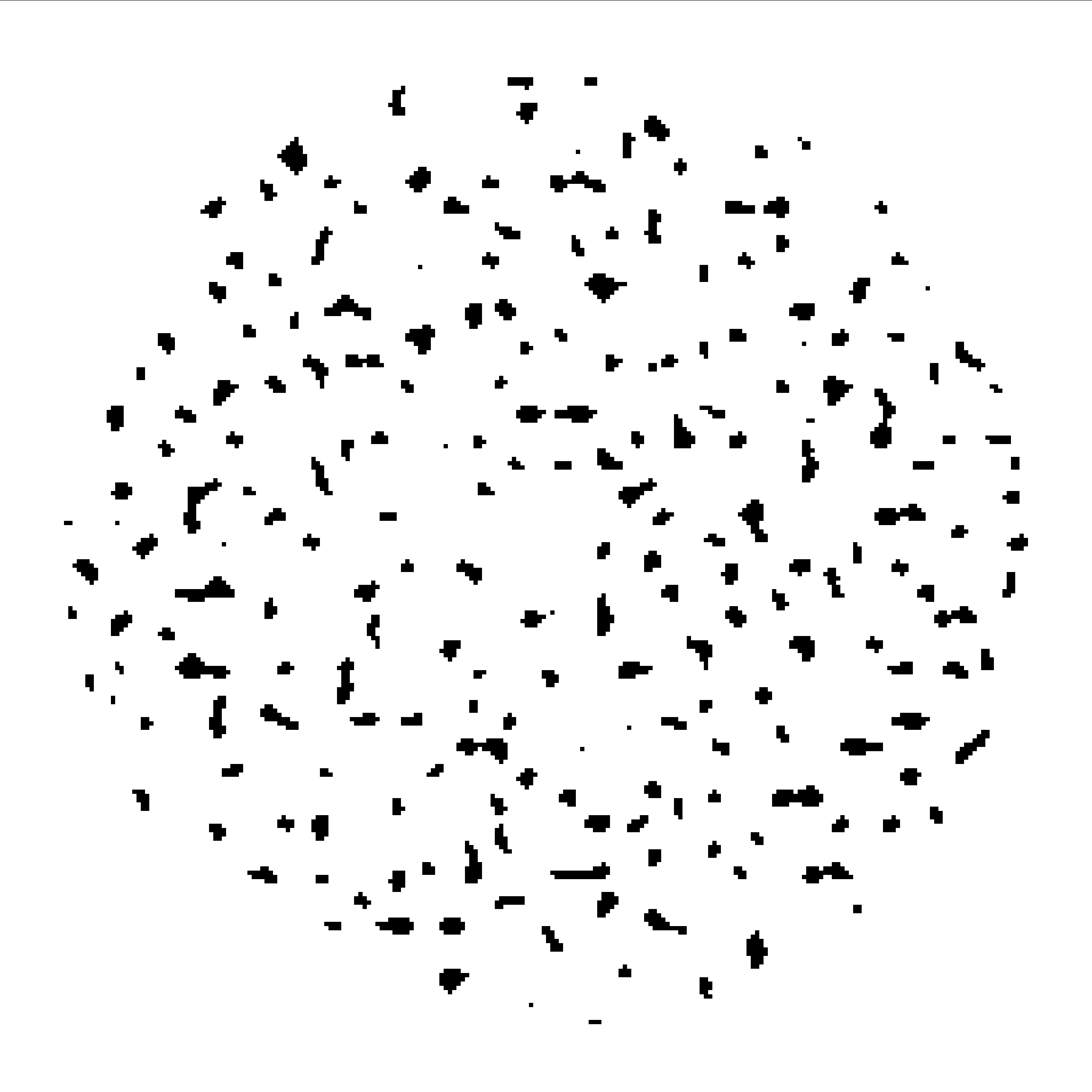}
	\includegraphics[width=.2\linewidth,trim={0cm 0cm 0cm 2cm},clip]{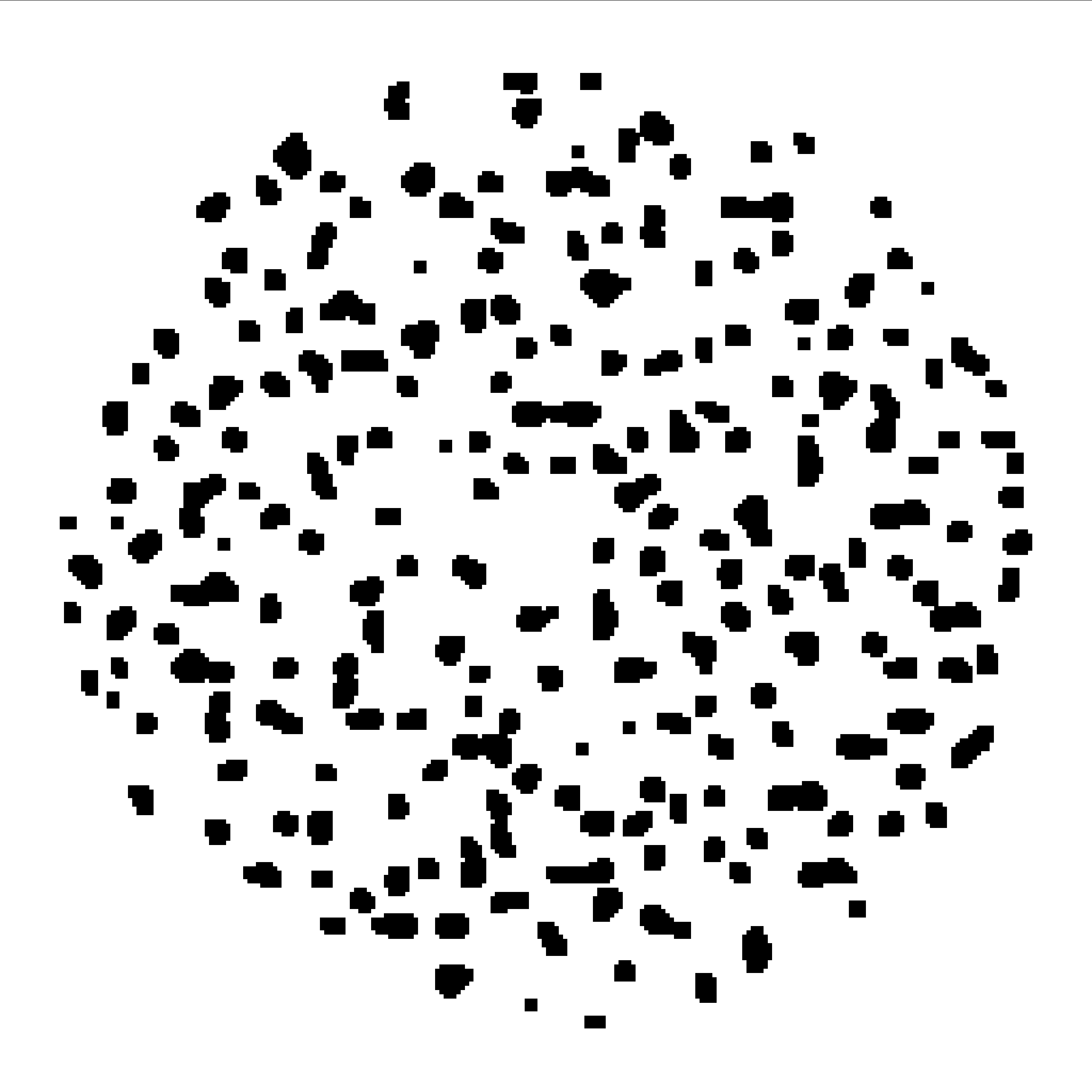}
	\includegraphics[width=.2\linewidth,trim={0cm 0cm 0cm 2cm},clip]{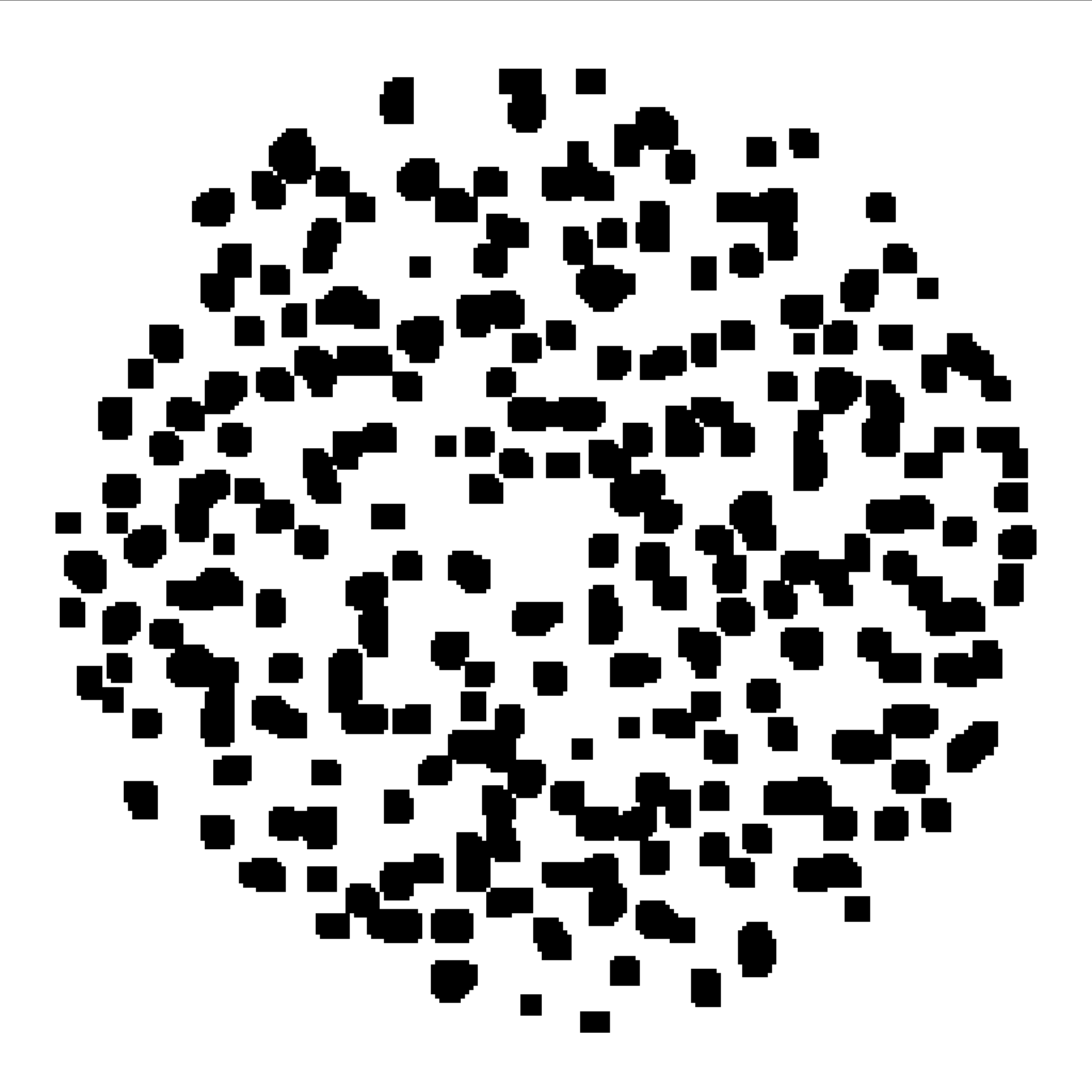}
	\includegraphics[width=.2\linewidth,trim={0cm 0cm 0cm 2cm},clip]{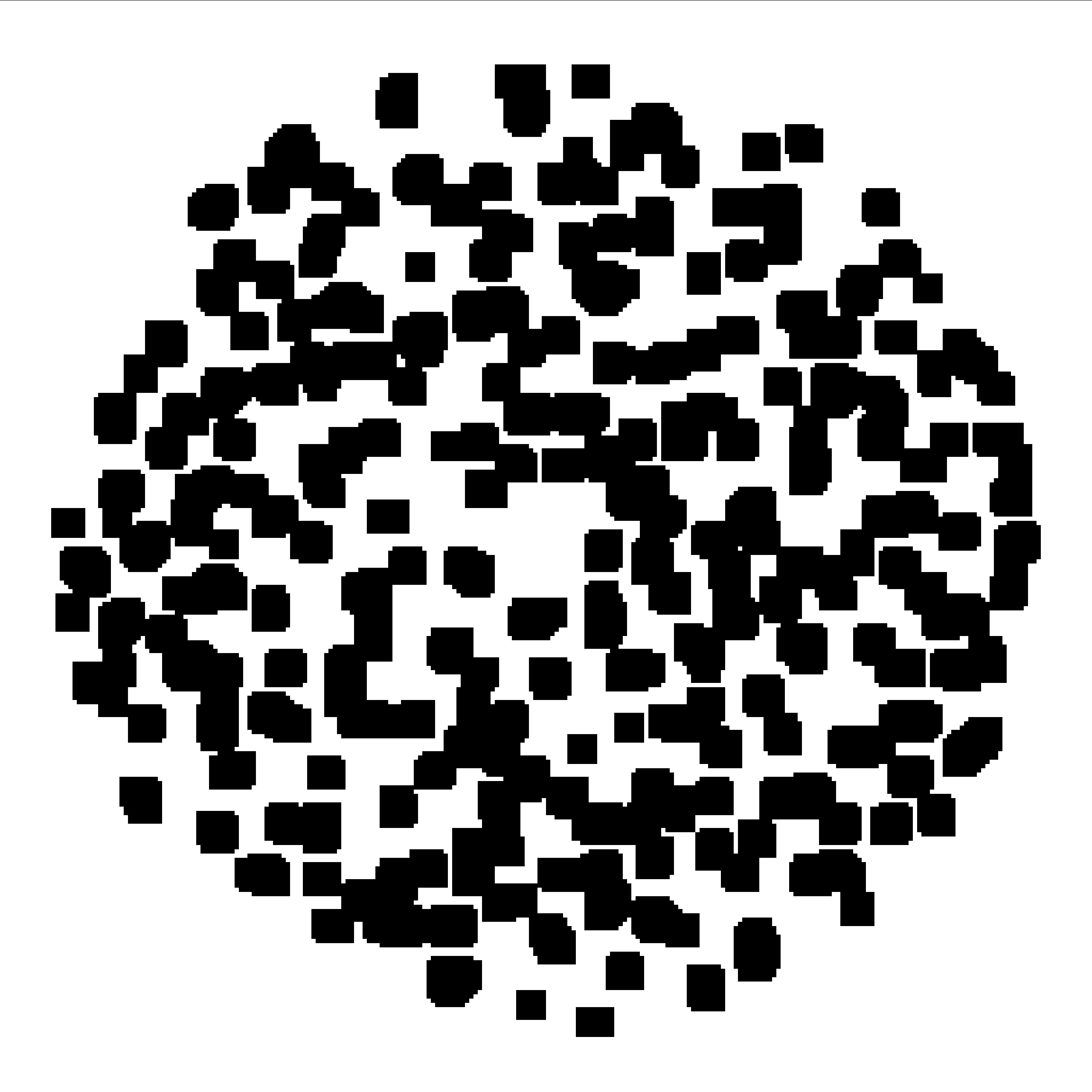}
	\vspace{-.4cm}
	\caption{\footnotesize{A map with obstacle expansion radii from left to right: 0.0m, 0.4m, 0.8m, 1.2m.}}
	\label{fig:cSpaceExpandedMaps}
\end{figure}

\vspace{-.3cm}
The obstacle expansion results shown in Figure \ref{fig:cSpaceData} illustrate the opposite trend compared to the results of the lookahead experiments discussed above.
This inverted trend is because, while expanding obstacle footprints led to fewer NR collisions, the maps became very difficult to navigate with newly expanded obstacles closing off many corridors, especially in cluttered regions.
Figure \ref{fig:cSpaceExpandedMaps} shows one map from the ablation experiment map set with progressively expanded obstacle footprints.
In some cases, especially at the higher expansion radii, the robot is completely surrounded by obstacles which prevents planning entirely.
\vspace{-.5cm}
\vspace{-.6cm}

In the field experiments, NR was unable to consistently find solutions because of the obstacle distributions.
The ablation experiment supports this finding, but also shows the benefit of obstacle expansion in less cluttered environments.
The planning times and path durations increased for NR because of the decreasing availability of viable regions to explore.
The consistent trends in the data of all three mapping types demonstrate the lack of sensitivity to obstacle configuration, indicating a robustness of the algorithms.
\vspace{-.2cm}
\vspace{-.3cm}
\section{Conclusion}\label{sec:conclusion}
\vspace{-.3cm}
Motion planning systems that operate at, or beyond, the perception horizon should predict potential controller behavior in order to improve the safety of autonomous sytems.
We present three methods of incorporating stochastic controller behavior into KEASL, and present experiments to evaluate the performance with varying levels of pure-pursuit lookahead and obstacle inflation parameters.
Although PER and GER were unable to find solutions within the 1-second time limit, the addition of a graph revision step in GEGR vastly improved the efficiency of search, and showed comparable run-times to NR.
The results show that simulating controller rollouts in the search process leads to more conservative plans with less likelihood for collisions, while maintaining comparable path durations and planning times over a normal search space.
\vspace{-.5cm}
\section*{Acknowledgements}\label{sec:acknowledgements}
\vspace{-.4cm}
\footnotesize{
Research was sponsored by the DEVCOM Army Research Laboratory (ARL) and was accomplished under Cooperative Agreement Number W911NF-20-2-0106 and W911NF-24-2-0227. The views and conclusions contained in this document are those of the authors and should not be interpreted as representing the official policies, either expressed or implied, of the Army Research Laboratory or the U.S. Government. The U.S. Government is authorized to reproduce and distribute reprints for Government purposes notwithstanding any copyright notation herein.
}
 
\bibliographystyle{splncs03}
\bibliography{root}

\end{document}